\documentclass[letterpaper,10pt,journal,twoside]{IEEEtran}

\usepackage{cite}
\usepackage[utf8]{inputenc}
\usepackage{graphicx}
\usepackage{amsmath}
\usepackage{amssymb}
\usepackage{booktabs}
\usepackage{array}
\usepackage{multirow}
\usepackage{tabularx}
\usepackage[caption=false,font=footnotesize]{subfig}
\usepackage{url}
\usepackage[hidelinks]{hyperref}
\usepackage{xcolor}

\definecolor{PEAfowlCyan}{HTML}{00A4B4}

\newcommand{\best}[1]{\textbf{#1}}
\newcommand{\second}[1]{\underline{#1}}

\newcommand{\rev}[1]{#1}
\newcommand{\revtext}[1]{#1}

\ifdefined\pdfminorversion
\fi

\newcommand{\RALAuthorList}{
Qingyu Fan$^{1}$,
Zhaoxiang Li$^{2,*}$,
Jinrui Hu$^{1,*}$,
Yi Lu$^{2}$,
Wang Chen$^{2}$,
Qiu Shen$^{2}$,
Xiao-xiao Long$^{2,\dagger}$,
Yinghao Cai$^{1,\dagger}$,
Tao Lu$^{1}$,
Shuo Wang$^{1}$,
and Xun Cao$^{2}$
}
\newcommand{\RALReceivedDate}{April 13, 2026}
\newcommand{\RALRevisedDate}{July 10, 2026}
\newcommand{\RALAcceptedDate}{August 11, 2026}
\newcommand{\RALEditorName}{Julia Borras Sol}
\newcommand{\RALFundingStatement}{This work was supported in part by New Generation Artificial Intelligence National Science and Technology Major
Project 2025ZD0122904  and National Science and Technology Major Project 2026ZD1609003,  and in part by National Natural Science Foundation of China (Grant Number:62473366, 62273342 and U23B2038).}
\newcommand{\RALAffiliationFootnotes}{%
\thanks{$^{1}$Qingyu Fan, Jinrui Hu, Yinghao Cai, Tao Lu, and Shuo Wang
are with the Institute of Automation, Chinese Academy of Sciences,
Beijing 100190, China
(e-mail: qingyu.fan.ai@outlook.com; yinghao.cai@ia.ac.cn).}%
\thanks{$^{2}$Zhaoxiang Li, Yi Lu, Wang Chen, Qiu Shen,
Xiao-xiao Long, and Xun Cao are with Nanjing University,
Nanjing 210023, China
(e-mail: xxlong@nju.edu.cn).}%
\thanks{$^{*}$Zhaoxiang Li and Jinrui Hu contributed equally to this work.}%
\thanks{$^{\dagger}$Corresponding authors: Xiao-xiao Long and Yinghao Cai.}%
}
\newcommand{\RALKeywords}{
Bimanual Manipulation,
Deep Learning in Grasping and Manipulation,
RGB-D Perception,
Vision-Language-Action Models
}
\title{PEAfowl: Perception-Enhanced Multi-View Vision-Language-Action for Bimanual Manipulation}

\author{\RALAuthorList%
\thanks{Manuscript received: \RALReceivedDate; Revised: \RALRevisedDate; Accepted: \RALAcceptedDate.}%
\thanks{This paper was recommended for publication by Editor \RALEditorName\ upon evaluation of the Associate Editor and Reviewers' comments. \RALFundingStatement}%
\RALAffiliationFootnotes%
\thanks{Author-accepted manuscript. Early Access citation: Q.~Fan \textit{et al.}, \textquotedblleft{}PEAfowl: Perception-Enhanced Multi-View Vision-Language-Action for Bimanual Manipulation,\textquotedblright{} \textit{IEEE Robotics and Automation Letters}, early access, 2026, doi: 10.1109/LRA.2026.3726379. Available: \url{https://ieeexplore.ieee.org/document/11661761}.}%
}

\begin{document}
\maketitle

\begin{abstract}

Bimanual manipulation in cluttered scenes requires policies that remain stable under occlusions, viewpoint changes and scene variations. Existing vision-language-action models often lack such robustness because (i) multi-view features are fused via view-agnostic token concatenation, yielding limited cross-view spatial representations, and (ii) language is injected as global conditioning, resulting in coarse instruction grounding.
In this paper, we introduce PEAfowl, a perception-enhanced multi-view VLA policy for bimanual manipulation. For spatial perception, PEAfowl predicts per-token depth distributions, performs differentiable 3D lifting, and aggregates local cross-view neighbors to form geometrically grounded, cross-view aligned representations. For language utilization, we propose to replace global conditioning with a Perceiver-style text-aware readout over frozen CLIP visual features, enabling iterative evidence accumulation. To better exploit commodity RGB-D sensing despite noisy and incomplete depth, PEAfowl's depth-distribution lifting naturally supports training-only depth distillation, where a pretrained depth teacher supervises the depth-distribution head to inject refined geometric priors without adding inference overhead.
On RoboTwin 2.0 under domain-randomized setting, PEAfowl improves the strongest baseline by 23.0 pp in success rate, and physical experiments further demonstrate improved performance on the evaluated real-robot tasks. 
Project website: {\color{PEAfowlCyan}\url{https://peafowlvla.github.io/}}.

\end{abstract}


\section{INTRODUCTION}

\IEEEPARstart{M}{ulti-task} bimanual manipulation requires a unified robot policy that integrates vision, language, and proprioception to follow diverse instructions under changing scenes and viewpoints. Recent vision-language-action (VLA) models have shown encouraging performance in single-arm manipulation \cite{brohan2023rt2,pmlr-v270-kim25c}. By leveraging large-scale robot data and web-scale vision-language supervision, a single policy can now execute a wide range of manipulation skills from natural-language prompts \rev{showing promising performance in both simulation and real-world environments}.

\begin{figure}[t]
  \centering
  \includegraphics[width=\linewidth,clip,trim=10pt 10pt 10pt 10pt]{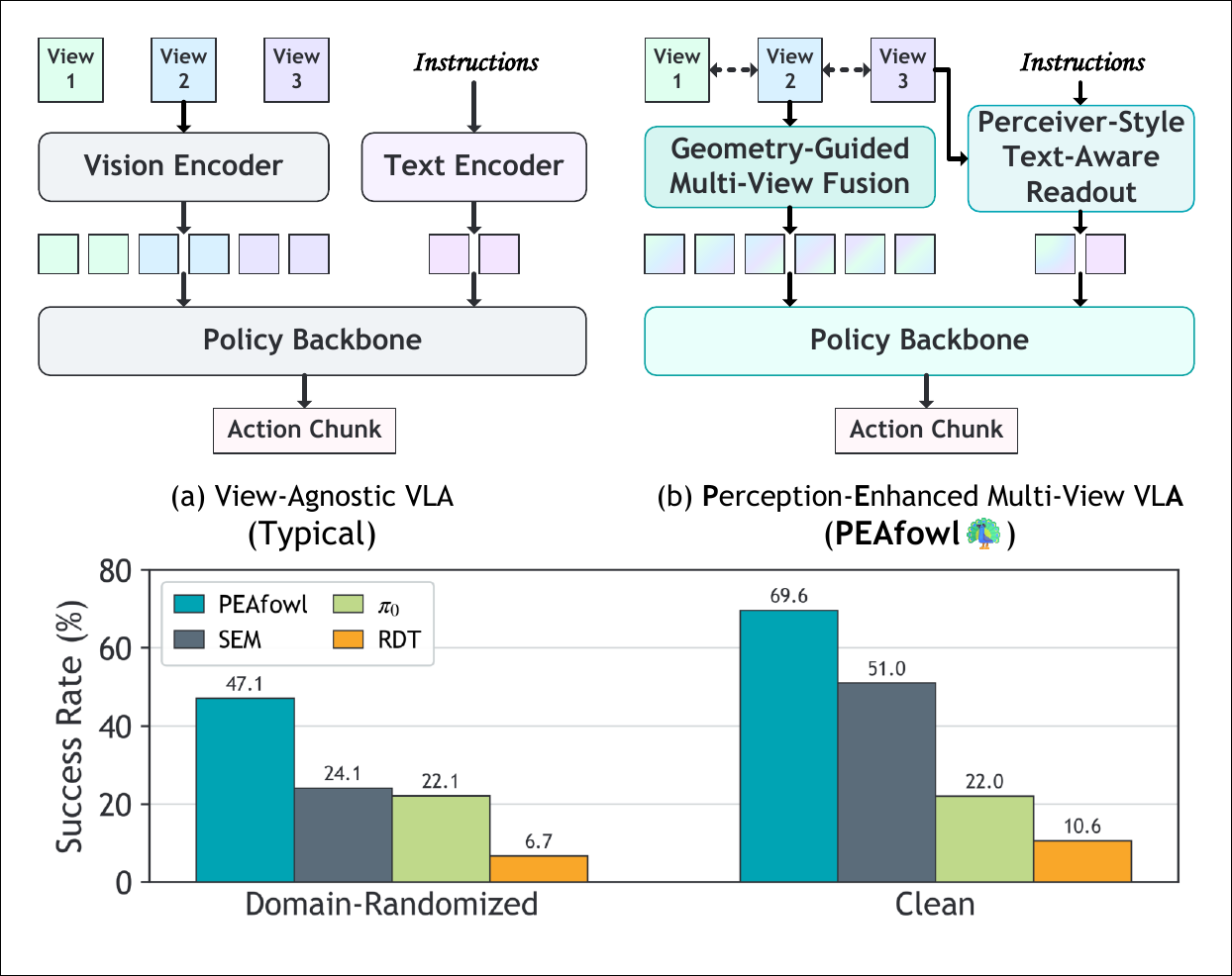}
  \caption{\textbf{Motivation and overview of PEAfowl.}
    (a) Prior bimanual VLAs typically concatenate per-view visual tokens and apply global text conditioning, without explicit cross-view geometric alignment or instruction-relevant visual evidence retrieval and aggregation. (b) We propose PEAfowl, which incorporates geometry-guided multi-view fusion and a Perceiver-style text-as-query readout over frozen CLIP features. Bottom: Average success rates on RoboTwin 2.0 (nine training tasks) under Clean and Domain-Randomized settings, compared with multi-task bimanual baselines.}
  \label{fig:fig1}
\end{figure}

However, multi-task bimanual manipulation remains underexplored, especially under challenging scene variations. Pretrained VLA policies often exhibit substantial performance degradation in cluttered scenes with distractors and under appearance and viewpoint variations. Generalization to unseen objects or scenes becomes particularly unstable for tasks that require coordinated bimanual motion \cite{chen2025robotwin}. The high-dimensional and tightly coupled bimanual action space increases the data requirements for learning \rev{robust} multi-task policies. 
Moreover, bimanual manipulation often involves frequent self-occlusions and inter-object occlusions as well as fine-grained language instructions, \rev{which place higher demands on spatial perception and instruction-conditioned visual readout. Under limited demonstrations, improving perception efficiency and instruction-relevant evidence aggregation is therefore critical for multi-task learning and cross-scene robustness.}

Bimanual manipulation platforms are inherently multi-view perception systems: multiple cameras are commonly deployed to increase workspace coverage and mitigate occlusions. These complementary views capture rich 3D geometric information that, when properly leveraged, can stabilize spatial understanding across scenes and camera configurations. However, most existing VLA pipelines treat multi-view inputs in a view-agnostic manner: each image is encoded independently, and tokens from different cameras are concatenated or stacked before being passed to a shared policy head \cite{pmlr-v270-kim25c,zhou2025sem,BlackK-RSS-25}. Such designs do not explicitly model cross-view geometric correspondences or \rev{encourage 3D-aligned feature interaction, making the learned representations more sensitive to changes in camera poses, calibration errors, and occlusions. }

Another key limitation lies in how language interacts with visual features. In many VLA architectures~\cite{pmlr-v270-kim25c,liu2025rdt}, language is injected as a global conditioning vector or via a small number of text tokens appended to the visual stream, while the dominant attention computation remains vision-centric. This strategy is often sufficient for low-clutter tasks, but in multi-task, multi-object scenes it may produce unfocused, instruction-agnostic attention and weak exploitation of pretrained vision-language alignment to identify relevant objects and spatial relations in the scene. Recent work suggests that text-aware visual feature extraction, in which language tokens explicitly query and aggregate relevant visual evidence, can improve grounding performance \cite{huang2025otter}, yet such strategies have not been explored for multi-view bimanual manipulation.

In this paper, we introduce PEAfowl, a multi-view vision-language-action model with geometry- and language-guided perception for bimanual manipulation. On the spatial side, we propose a geometry-driven multi-view fusion module that (i) performs per-patch RGB-D token fusion, (ii) predicts a depth distribution for each visual token, and (iii) backprojects tokens into a shared 3D base frame to perform local 3D neighborhood aggregation across cameras. This design explicitly models cross-view geometric correspondences and endows 2D visual features with depth-aware and \rev{cross-view-aligned} structure.

On the language side, PEAfowl builds upon OTTER-style text-aware visual extraction \cite{huang2025otter}, and replaces global text conditioning with a Perceiver-style text-aware transformer \cite{jaegle2021perceiverio}. Text tokens act as latent queries that iteratively cross-attend to per-view patch features, producing a compact set of language-conditioned visual tokens. This mechanism sharpens attention over task-relevant objects and spatial relations suitable for policy learning. 
Fig.~\ref{fig:fig1} provides an overview of this architecture and contrasts it with a standard VLA pipeline.

\rev{Commodity RGB-D sensors provide depth measurements for real-robot manipulation, but these measurements are often noisy and incomplete. Since PEAfowl represents geometry through per-token depth distributions, completed depth can be used as a training-only supervisory signal for 3D lifting.}
We leverage a pretrained Camera Depth Model (CDM) \cite{liu2025manipulation} as a training-time depth teacher while still feeding only raw depth images to the policy at both training and test time, which introduces no test-time overhead and transfers refined geometric priors into the multi-view policy.

Both simulation and physical experiments demonstrate that PEAfowl, despite having only \rev{\textbf{312M}} trainable parameters, outperforms existing bimanual VLA models and visuomotor baselines, and \rev{shows stronger robustness} to novel scene appearances, workspace geometries, and language instructions. In simulation, we evaluate on $9$ tasks from RoboTwin 2.0 \cite{chen2025robotwin} under both clean and heavily domain-randomized settings. On physical platforms, PEAfowl demonstrates \rev{improved performance}, particularly on tasks requiring accurate bimanual coordination. To summarize, our contributions are:

\begin{enumerate}
    \item  We introduce a geometry-guided multi-view perception module that builds spatially-aware representations from multi-view RGB-D observations, along with the depth distillation scheme for real-world deployment.
        
    \item We propose to replace common global text conditioning in visual-language interaction with a Perceiver-style text-as-query readout over frozen CLIP features, which produces compact instruction-conditioned tokens.
    \item Extensive experiments in both simulation and real-world bimanual manipulation tasks demonstrate the benefits of geometry- and language-guided multi-view perception.

\end{enumerate}

\begin{figure*}[t]
  \centering
  \includegraphics[width=\linewidth,clip,trim=20pt 10pt 20pt 20pt]{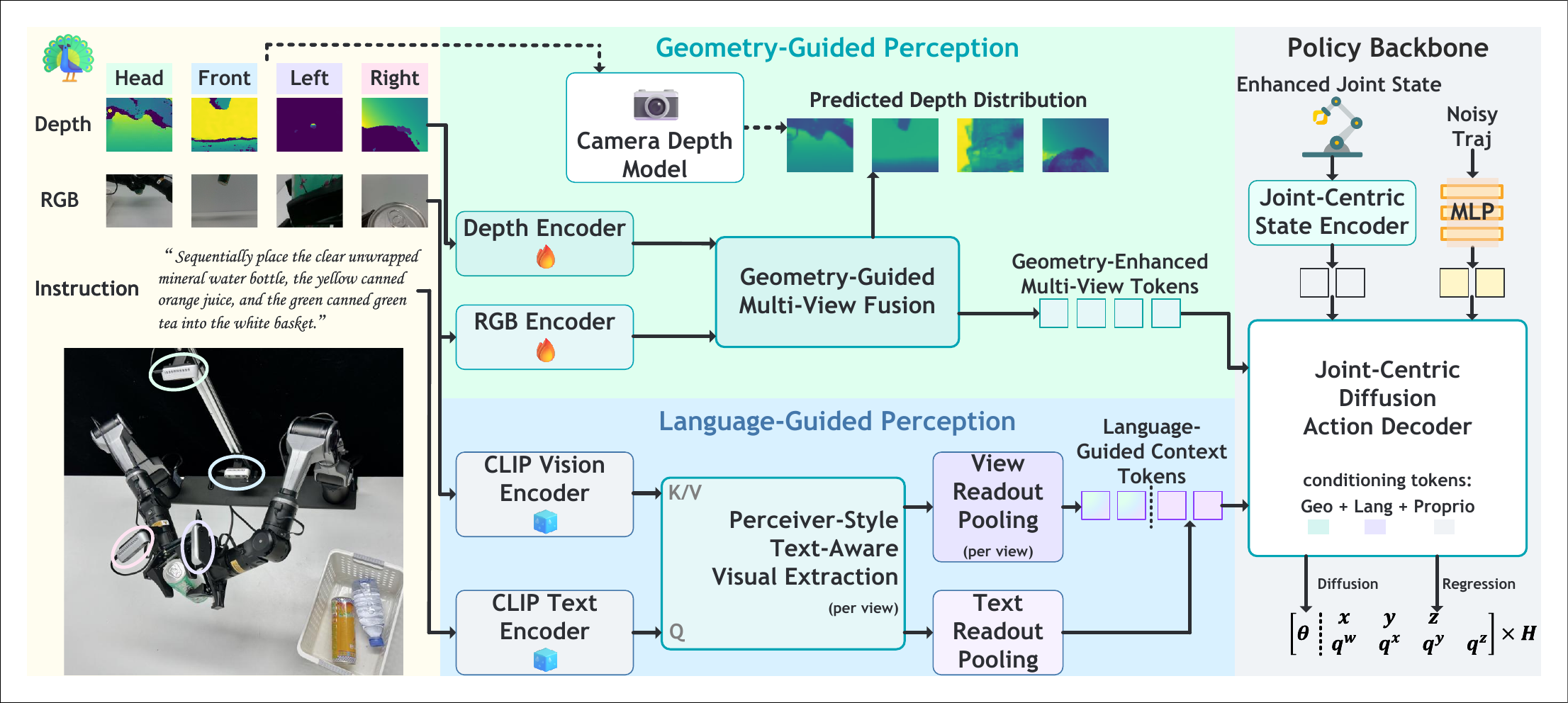}
  \caption{\textbf{PEAfowl architecture. }
  PEAfowl couples geometry-guided multi-view fusion with language-guided readout to condition a SEM-style diffusion action decoder. Top: RGB–D tokens are used to predict per-token depth distributions for differentiable 3D lifting and cross-view fusion; a pretrained camera depth model supervises the depth-distribution head during training only. Bottom: Frozen CLIP features are queried by a Perceiver-style text-as-query readout and pooled into compact context tokens.}
  \label{fig:fig2}
\end{figure*}

\section{RELATED WORK}

\subsection{Multi-view Perception}

Multi-view perception has been studied extensively in classical structure-from-motion and multi-view stereo, and more recently in neural reconstruction and multi-view transformers~\cite{wang2025vggt}. Most of these approaches are typically optimized for offline 3D perception  and rely on heavy backbones or dense geometric supervision, which limits their suitability as lightweight perception modules tightly coupled with robot policy learning.

In robotic manipulation, multi-view observations are used to improve robustness to viewpoint variations and mitigate occlusions~\cite{seita2023mvmwm,li2025threedmvp,Goyal-RSS-24,fan2025neugrasp}. While these methods demonstrate the benefit of multi-view sensing, many collapse views into view-invariant embeddings or fuse per-view features through simple concatenation, where cross-view 3D consistency is not explicitly considered and depth uncertainty is largely ignored. In contrast, PEAfowl introduces a geometry-guided multi-view perception module that maintains cross-view spatial tokens under noisy depth measurements. This policy-friendly multi-view perception design explicitly leverages geometry under noisy RGB-D observations, improving robustness to viewpoint changes and scene variations.

\subsection{Vision-Language-Action Models}

VLA policies map visual observations and language instructions to actions~\cite{brohan2023rt2}. Existing methods and datasets largely focus on single-arm manipulation. In multi-view settings, images are commonly treated as view-agnostic 2D tokens, with language injected as global conditioning. Recent bimanual VLAs still rely primarily on 2D RGB features and multi-image concatenation~\cite{liu2025rdt,BlackK-RSS-25}. Spatially enhanced VLAs instead inject explicit 3D information into policy representations~\cite{QuD-RSS-25}. SEM aligns multi-view features and robot states in a shared base frame to form spatial representations~\cite{zhou2025sem}. For language conditioning, OTTER improves grounding via computing similarity scores between each instruction token and each visual feature to highlight the visual information relevant to the instruction~\cite{huang2025otter}. PEAfowl combines 3D-aligned multi-view tokens with iterative text-as-query readout to enable spatial reasoning and more accurate grounding in cluttered bimanual scenes. This helps the policy more reliably disambiguate instruction-relevant targets in cluttered scenes and under scene variations.

\section{METHOD}

Fig.~\ref{fig:fig2} presents an overview of PEAfowl. The framework integrates geometry-guided multi-view fusion with language-guided readout to produce policy-ready tokens for a joint-centric diffusion action decoder.

\subsection{Problem Formulation}

Let $\mathcal{V}=\{1,\dots,V\}$ denote the camera set. The robot receives multi-view RGB-D observations $\mathbf{o}_t \triangleq \{\mathbf{I}^{(v)}_t, \mathbf{D}^{(v)}_t\}_{v\in\mathcal{V}}$, along with proprioceptive state $\mathbf{s}_t\in\mathbb{R}^{d_s}$ and instruction $\ell$. Our goal is to learn a bimanual policy $\pi_\phi$ that predicts an action chunk of horizon $H$ conditioned on $(\mathbf{o}_t,\ell,\mathbf{s}_t)$:
\begin{equation}
\hat{\mathbf{a}}_{t:t+H-1} = \pi_\phi(\mathbf{o}_t, \ell, \mathbf{s}_t)
\label{eq:policy_chunk}
\end{equation}
where $\mathbf{a}_t = [\mathbf{a}^L_t;\mathbf{a}^R_t]\in\mathbb{R}^{d_a}$ concatenates the left- and right-arm control commands. 

\subsection{Geometry-Guided Multi-View Fusion}
\label{sec:geo_fusion}

\begin{figure}[t]
  \centering
  \includegraphics[width=\linewidth,clip,trim=10pt 10pt 10pt 10pt]{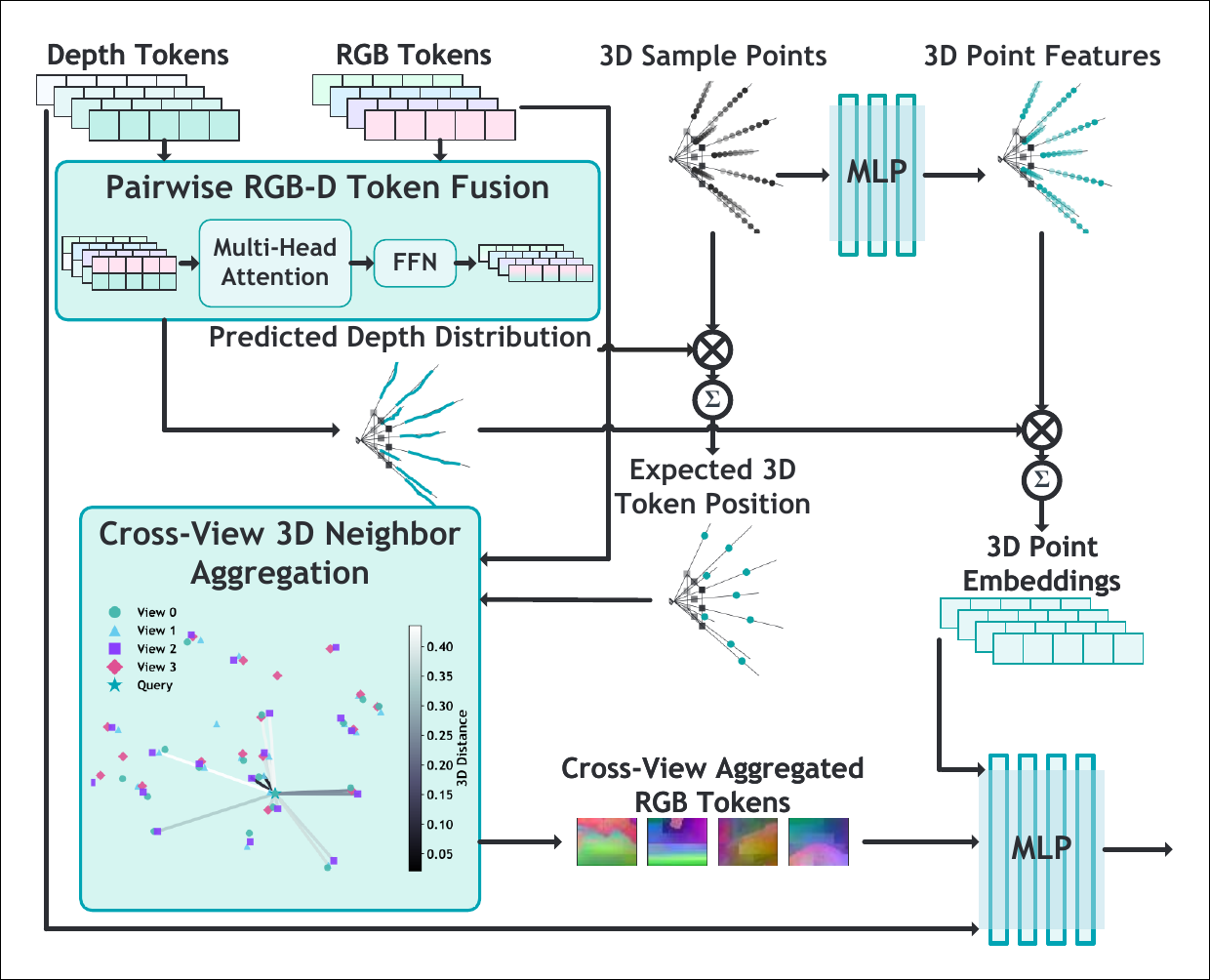}
  \caption{\textbf{Geometry-Guided Multi-View Fusion (GGMVF).}
  Multi-scale RGB and depth features are tokenized, and co-located RGB-D pairs are used to predict discrete depth distributions for differentiable 3D lifting. The resulting 3D anchors enable top-$K$ cross-view neighbor aggregation in the base frame using distance-based softmax weights and a gated residual update. Aggregated RGB tokens are fused with depth tokens and 3D point embeddings via an MLP to produce geometry-enhanced tokens.}
  
  \label{fig:fig3}
\end{figure}

Bimanual manipulation in cluttered scenes can be easily affected by self-occlusions and inter-object occlusions, which motivates explicit 3D alignment for multi-view fusion. Given multi-view RGB-D pyramids, our geometry-guided fusion produces \rev{cross-view-aligned tokens} via (i) per-patch RGB-D fusion, (ii) per-token depth distributions for differentiable 3D lifting, and (iii) cross-view neighbor aggregation in a shared base frame as shown in Fig.~\ref{fig:fig3}.

\paragraph{Multi-view RGB-D Feature Extraction}

For each view $v$, we extract multi-scale RGB and depth feature pyramids with modality-specific encoders shared across cameras. The RGB encoder is initialized from Grounding-DINO \cite{liu2024groundingdino} and depth features are extracted using a lightweight ResNet \cite{he2016resnet} to obtain $\{\mathbf{F}^{(v)}_{\mathrm{rgb},l}\}_{l=1}^{L}$ and $\{\mathbf{F}^{(v)}_{\mathrm{dep},l}\}_{l=1}^{L}$.

\paragraph{Tokenization across Scales}

We flatten the $L$-level feature pyramids into RGB and depth token sequences, denoted by
$\mathbf{T}^{(v)}_{\mathrm{rgb}} = \{\,\mathbf{t}^{(v)}_{\mathrm{rgb},n}\,\}$
and $\mathbf{T}^{(v)}_{\mathrm{dep}} = \{\,\mathbf{t}^{(v)}_{\mathrm{dep},n}\,\}$.
Known camera intrinsics and extrinsics are used for backprojection in the subsequent 3D lifting.

\paragraph{Depth-Aware 3D Lifting}

To model depth ambiguity at the patch level, we predict a discrete depth distribution $\mathbf{p}^{(v)}_n$ over $B$ bins from co-located RGB–D token pairs, and use it to softly lift each 2D token into the robot base frame. For view $v$ and token $n$, the token center $\mathbf{u}^{(v)}_n$ is backprojected at each depth bin $d_i$ to obtain 3D samples $\mathbf{x}^{(v)}_{n,i}$. We then compute the expected 3D anchor and a depth-aware point embedding by distribution-weighted aggregation:
\begin{equation}
\bar{\mathbf{x}}^{(v)}_n = \sum_{i=1}^{B} p^{(v)}_{n,i}\,\mathbf{x}^{(v)}_{n,i},
\qquad
\mathbf{g}^{(v)}_n = \sum_{i=1}^{B} p^{(v)}_{n,i}\,\phi(\mathbf{x}^{(v)}_{n,i})
\label{eq:depth_lifting}
\end{equation}
where $\phi(\cdot)$ projects 3D coordinates to point features. We use $\bar{\mathbf{x}}^{(v)}_n$ for cross-view matching and $\mathbf{g}^{(v)}_n$ as the geometric context for subsequent fusion.

\paragraph{Pairwise RGB-D Token Fusion}
To incorporate noisy commodity depth measurements \rev{while preserving} pretrained RGB semantics, we apply a lightweight \emph{local} fusion only on co-located RGB-D token pairs and only in the depth-distribution branch.
The RGB token is projected to the depth feature dimension using a linear transformation matrix 
$\mathbf{W}_r$ and fused with the length-2 pair using a small attention block: \begin{equation}
[\hat{\mathbf{r}}^{(v)}_n,\hat{\mathbf{d}}^{(v)}_n]
=
\mathrm{PairAttn}\big([\mathbf{W}_r\mathbf{t}^{(v)}_{\mathrm{rgb},n},\,\mathbf{t}^{(v)}_{\mathrm{dep},n}]\big)
\label{eq:pairwise_rgbd}
\end{equation}

By restricting attention to co-located RGB-D pairs, the fusion learns \rev{selective} mixing while leaving the main RGB stream unchanged.

\paragraph{Cross-View 3D Neighbor Aggregation}

Given expected 3D token anchors $\{\bar{\mathbf{x}}^{(v)}_n\}$ from Eq.~\eqref{eq:depth_lifting}, we aggregate information across views based on geometric proximity in the shared base frame. For each query token $(v,n)$, we compute pairwise distances $\delta^{(v,w)}_{n,m}=\|\bar{\mathbf{x}}^{(v)}_n-\bar{\mathbf{x}}^{(w)}_m\|_2$ to tokens $(w,m)$ from other views, select the $K$ nearest neighbors $\mathcal{N}^{(v)}_n$, and aggregate their RGB tokens with a distance-softmax:
\begin{equation}
\alpha^{(v,w)}_{n,m}
=
\frac{\exp(-\delta^{(v,w)}_{n,m}/\tau)}{\sum_{(w',m')\in\mathcal{N}^{(v)}_n}\exp(-\delta^{(v,w')}_{n,m'}/\tau)}
\label{eq:mv_agg1}
\end{equation}
where $\tau$ controls the softness of geometric matching. The aggregated feature is then computed as:
\begin{equation}
\mathbf{h}^{(v)}_n
=
\sum_{(w,m)\in\mathcal{N}^{(v)}_n}\alpha^{(v,w)}_{n,m}\,\mathbf{t}^{(w)}_{\mathrm{rgb},m}
\label{eq:mv_agg2}
\end{equation}
The aggregated feature is then added to the original token via a gated residual update:\begin{equation}
\tilde{\mathbf{t}}^{(v)}_{\mathrm{rgb},n}=\mathbf{t}^{(v)}_{\mathrm{rgb},n}+\gamma\,\mathbf{h}^{(v)}_n
\end{equation}
where $\gamma$ is a learnable scalar gate. This geometry-based aggregation aligns tokens that correspond to the same physical region across cameras, thereby improving robustness to occlusions and viewpoint changes. 
The resulting tokens are fused with depth tokens and 3D point embeddings via an MLP to form inputs to the action decoder.

\subsection{Language-Guided Multi-View Readout}
\label{sec:lang_readout}

To improve instruction \rev{alignment} in cluttered multi-object scenes, we replace global text conditioning with a text-aware multi-view readout over frozen CLIP features. For each view, we use a Perceiver-style extractor that iteratively updates text latents by cross-attending to CLIP patch tokens, yielding compact instruction-conditioned tokens for the policy.

\paragraph{Frozen CLIP Tokens}
Given instruction $\ell$ and multi-view RGB images $\{\mathbf{I}^{(v)}\}_{v\in\mathcal{V}}$, we extract frozen CLIP text tokens and per-view patch tokens \cite{radford2021clip}. For denser vision-language alignment, we use the final attention-layer outputs of the CLIP vision encoder~\cite{zhu2024clearclip}:
\begin{equation}
\mathbf{T}_{\mathrm{txt}}=\mathrm{CLIP}_{\mathrm{txt}}(\ell)\in\mathbb{R}^{K_{\mathrm{txt}}\times D_c}
\end{equation}
\begin{equation}
\mathbf{X}^{(v)}=\mathrm{CLIP}_{\mathrm{img}}^{\mathrm{attn\mbox{-}last}}(\mathbf{I}^{(v)})\in\mathbb{R}^{N_p\times D_c}
\end{equation}

\paragraph{Perceiver-Style Text-as-Query Readout}
For each view $v$, latent queries are initialized with text tokens and refined through $M$ latent blocks that alternate cross-attention to patches and latent self-attention:
\begin{equation}
\mathbf{Z}^{(v,0)}=\mathbf{T}_{\mathrm{txt}}
\end{equation}
\begin{equation}
\mathbf{Z}^{(v,m+1)}=\mathrm{LatentBlock}\!\left(\mathbf{Z}^{(v,m)},\,\mathbf{X}^{(v)}\right),m=0,..,M-1
\end{equation}
which produce \rev{vision-conditioned} text latents $\mathbf{Z}^{(v)}\triangleq\mathbf{Z}^{(v,M)}\in\mathbb{R}^{K_{\mathrm{txt}}\times D_c}$. ReZero-gated residuals are used for stable stacking~\cite{bachlechner2021rezero}.

\paragraph{Readout Tokens}
We summarize tokens into a fixed number of context tokens via attention pooling:
\begin{equation}
\mathbf{R}^{(v)}=\mathrm{Pool}\!\left(\mathbf{Z}^{(v)}\right)\in\mathbb{R}^{R\times d},\
\mathbf{R}_{\mathrm{txt}}=\mathrm{Pool}\!\left(\mathbf{T}_{\mathrm{txt}}\right)\in\mathbb{R}^{R\times d}
\end{equation}
and concatenate them as the language-guided context sequence:
\begin{equation}
\mathbf{S}=[\mathbf{R}_{\mathrm{txt}};\mathbf{R}^{(1)};\dots;\mathbf{R}^{(V)}]\in\mathbb{R}^{(R+V\times R)\times d}
\end{equation}
which conditions the action decoder together with geometry-enhanced tokens (Sec.~\ref{sec:geo_fusion}) and proprioception.

\subsection{Policy Backbone}
\label{sec:policy}

\revtext{PEAfowl uses the SEM-style joint-centric policy backbone~\cite{zhou2025sem}.
At each time step, the robot state is represented by joint-centric tokens, where each token contains the joint angle and its forward-kinematics pose in the robot base frame.
A joint-centric diffusion transformer predicts an $H$-step bimanual trajectory conditioned on the encoded robot state, geometry-enhanced multi-view tokens, and language-guided context tokens.
Compared with SEM, PEAfowl keeps the joint-centric diffusion policy unchanged and modifies the perception modules before action decoding, including the geometry-guided multi-view fusion and language-guided readout described in Secs.~\ref{sec:geo_fusion} and~\ref{sec:lang_readout}.
}

\subsection{Training Objectives}
\label{sec:train_obj}

We optimize PEAfowl with a joint-centric diffusion imitation objective for policy learning and an auxiliary depth distillation loss for geometry-guided perception.

\paragraph{Diffusion Imitation Loss}
The diffusion decoder predicts an $H$-step bimanual joint trajectory. Given a noisy input, the decoder outputs the denoised sequence $\hat{\mathbf{J}}_{t:t+H-1}$. Training minimizes a weighted diffusion loss:
\begin{equation}
\mathcal{L}_{\text{diff}}
=
\left\|
(\hat{\mathbf{J}}-\mathbf{J})\,\mathbf{W}
\right\|_2^2
\label{eq:loss_diff}
\end{equation}
where $\mathbf{W}$ is a diagonal weight matrix. \revtext{Following SEM~\cite{zhou2025sem}, we add a forward-kinematics consistency term $\mathcal{L}_{\text{fk}}$ by recomputing joint poses from the predicted joint angles.}

\paragraph{Depth Distillation with Camera Depth Models}
\label{sec:depth_distill}
Our geometry-guided branch predicts a per-token depth distribution $\mathbf{p}^{(v)}_n$ (Sec.~\ref{sec:geo_fusion}), which is well suited for depth distillation from a pretrained depth teacher to inject geometry-aware priors into the policy.
We adopt the Camera Depth Model (CDM) from \cite{liu2025manipulation} and preprocess the training set offline to obtain refined depths
$\tilde{\mathbf{D}}^{(v)}=f_{\text{CDM}}(\mathbf{I}^{(v)},\mathbf{D}^{(v)})$,
from which we derive soft depth-bin targets $\mathbf{q}^{(v)}_n$.
Specifically, for each refined pixel depth $\tilde d$, we construct a soft target over the discretized depth bins by linearly interpolating to its two nearest bin centers, yielding a sparse two-hot distribution. These per-pixel targets are then average-pooled to token-level targets aligned with the multi-scale feature strides, producing $\mathbf{q}^{(v)}_n$ for supervising the per-token depth-distribution head. We further compute the validity weight $\omega^{(v)}_n$ as the fraction of valid pixels within each pooled region, so that unreliable or missing-depth regions contribute less to the supervision.
During both training and inference, the policy input remains the raw depth $\mathbf{D}^{(v)}$; \rev{the depth teacher is used only to produce completed-depth supervision labels during training and introduces no inference overhead}.
The depth distribution is supervised with a validity-weighted soft-label BCE loss:
\begin{equation}
\mathcal{L}_{\text{depth}}
=
\frac{1}{VN}\sum_{v\in\mathcal{V}}\sum_{n=1}^{N}
\omega^{(v)}_n\cdot
\mathrm{BCE}\!\left(\mathbf{p}^{(v)}_n,\mathbf{q}^{(v)}_n\right)
\label{eq:loss_depth}
\end{equation}

\paragraph{Total Loss}
\label{sec:total_loss}
The overall training objective combines diffusion imitation, FK consistency, and depth distillation:
\begin{equation}
\mathcal{L}
=
\mathcal{L}_{\text{diff}}
+
\lambda_{\text{fk}}\mathcal{L}_{\text{fk}}
+
\lambda_{\text{depth}}\mathcal{L}_{\text{depth}}
\label{eq:loss_total}
\end{equation}

\section{EXPERIMENTS}

\subsection{Experimental Setup}

\paragraph{Implementation Details}
\revtext{PEAfowl uses a single-step observation history and predicts an action chunk with horizon $H=64$, of which $32$ steps are executed in a receding-horizon manner. RGB-D inputs are resized to $320\times256$ from four camera views. The geometry branch uses feature strides $(8,16,32,64)$, while the diffusion decoder attends to the stride-16 and stride-32 levels. Per-token depth distributions are predicted over $B=128$ linearly spaced bins in $[0.01,1.2]$ m, and cross-view aggregation uses top-$K=16$ neighbors with distance-softmax temperature $\tau=0.08$ m. The pairwise RGB-D fusion block contains one four-head attention layer followed by an FFN with hidden dimension $128$. Training uses AdamW with learning rate $1.41\times10^{-4}$, weight decay $5\times10^{-4}$, batch size $16$, and $4\times10^5$ updates. Training uses a 1,000-step DDPM scheduler with squared-cosine beta schedule, while inference uses a DPMSolverMultistep scheduler with 10 denoising steps.}

\begin{table}[t]
  \centering
  \scriptsize
  \setlength{\tabcolsep}{2.4pt}
  \renewcommand{\arraystretch}{1.2}
  \caption{\revtext{\textbf{Same-hardware runtime analysis.}}
  Latency is measured on a single NVIDIA RTX 4090 with batch size one, four RGB-D views at $320\times256$, 10 warm-up iterations, 100 timed iterations, and 10 diffusion denoising steps.}
  \label{tab:runtime_analysis}
  \begin{tabular}{lcc}
    \toprule
    \textbf{Item} & \textbf{SEM} & \textbf{PEAfowl} \\
    \midrule
    Total params (M) & 183.9 & 740.0 \\
    Trainable params (M) & 183.9 & 312.4 \\
    Peak GPU memory (GB) & 0.95 & 3.04 \\
    Preprocess latency (ms) & 23.2 & 23.9 \\
    Model latency (ms) & 291.3 & 301.2 \\
    Total latency (ms) & 314.5 & 325.1 \\
    Throughput (Hz) & 3.18 & 3.08 \\
    \bottomrule
  \end{tabular}
\end{table}

\revtext{As shown in Tab.~\ref{tab:runtime_analysis}, PEAfowl increases total latency by 10.6 ms, corresponding to a 3.4\% overhead, while retaining a comparable policy-query frequency. The increased memory mainly comes from frozen CLIP encoders used by the language-guided readout.}

\paragraph{Simulation Setup}

We evaluate our method in simulation on RoboTwin~2.0 \cite{chen2025robotwin}, a scalable benchmark for bimanual manipulation. Two environment settings are used: Clean and Domain-Randomized (DR). The clean setting uses fixed visual and physical parameters with templated instructions for in-distribution evaluation, while the DR setting applies strong structured perturbations, e.g., distractors, lighting, texture, tabletop height, and instruction paraphrases, to stress-test robustness and \rev{cross-scene generalization}. All simulation experiments use the Aloha-AgileX embodiment with a 4-camera RGB-D setup. Unless otherwise specified, results are reported on $9$ tasks spanning short-, mid-, and long-horizon interactions, with $50$ demonstrations per task. VLA models are trained jointly across all tasks, while visuomotor policies are trained separately for each task.

\paragraph{Real-World Setup}
Experiments are conducted on a dual-arm AgileX Piper platform that mirrors the simulation configuration. Our real-robot setting includes $6$ tasks: $4$ adapted from simulation for sim-to-real evaluation, and $2$ real-only tasks included in the real-world training data. We collect $100$ high-quality demonstrations per task via VR teleoperation. Methods are initialized from simulation-pretrained models and then fine-tuned separately for each task on the corresponding real-world dataset. \revtext{The real-world protocol follows representative real-robot manipulation evaluations such as OpenVLA~\cite{pmlr-v270-kim25c} and DP3~\cite{Ze-RSS-24}, where each task is evaluated with 10 real-robot trials.}

\begin{figure}[t]
  \centering
  \includegraphics[width=\linewidth,clip,trim=10pt 10pt 10pt 10pt]{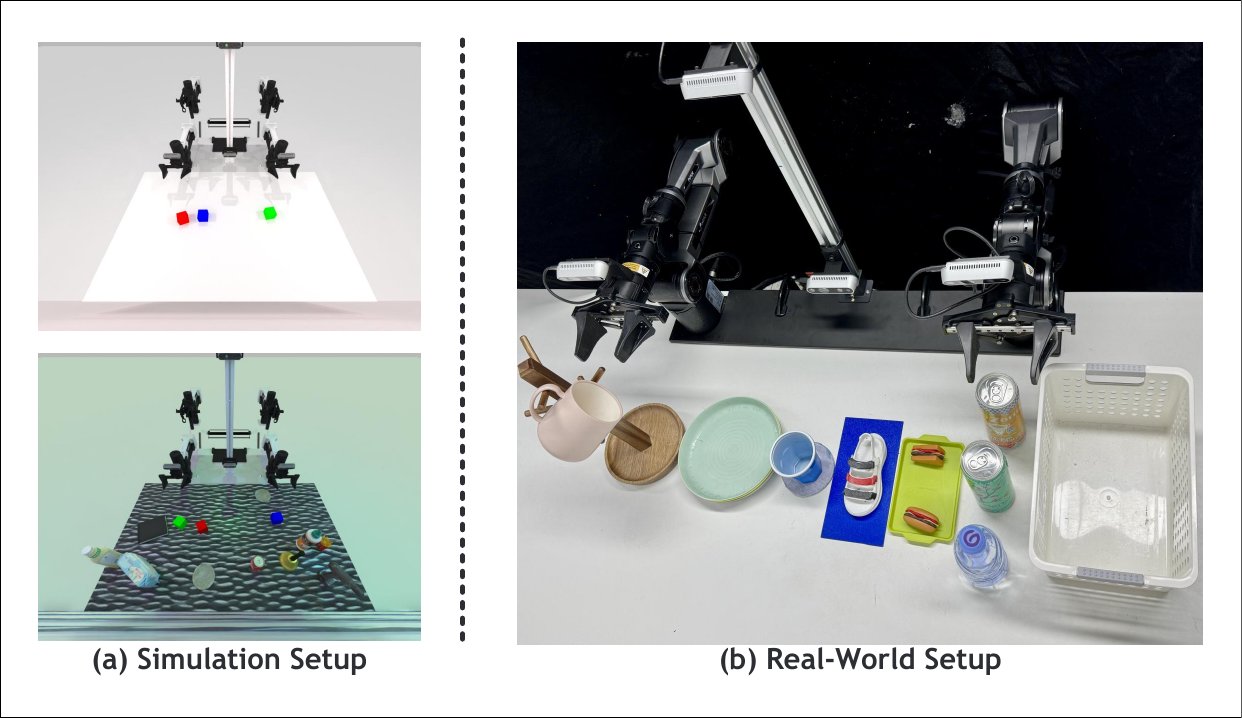}
  \caption{\textbf{Simulation and real-world setups.} (a) RoboTwin 2.0 simulation (Aloha-AgileX, 4-camera RGB-D) under Clean (top) and Domain-Randomized (bottom) settings. (b) Dual-arm AgileX Piper with a 4-camera rig.}
  \label{fig:fig4}
\end{figure}

We use four Intel RealSense D435 RGB-D cameras with aligned depth-to-color streams, each providing $424\times240$ RGB and depth at $30$ FPS, and camera extrinsics are provided by external rig calibration. During preprocessing, we apply a calibration-preserving remap to a fixed target resolution of $320\times256$ with a fixed destination intrinsic matrix. Compared with naive resizing, this remap effectively crops the usable source region, leading to a narrower field of view and reduced cross-view overlap. As a result, real-robot multi-view correspondence and fusion become more challenging. Fig.~\ref{fig:fig4} provides an overview of the real-world setup.

\begin{figure}[t]
  \centering
  \includegraphics[width=\linewidth,clip,trim=15pt 15pt 15pt 15pt]{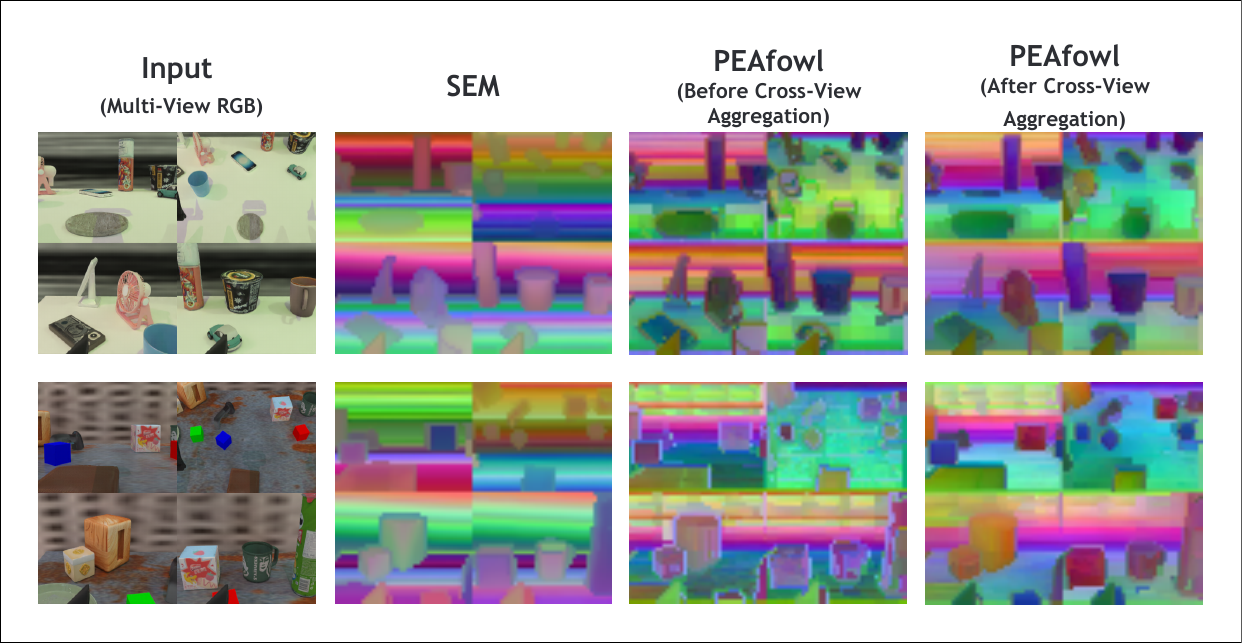}
  \caption{\textbf{Cross-view token consistency.} 
  Compared with SEM and PEAfowl pre-aggregation, PEAfowl post-aggregation yields more \revtext{coherent cross-view clusters} for the same physical regions across cameras.}
  \label{fig:fig5}
\end{figure}

\begin{figure}[t]
  \centering
  \includegraphics[width=\linewidth,clip,trim=15pt 15pt 15pt 15pt]{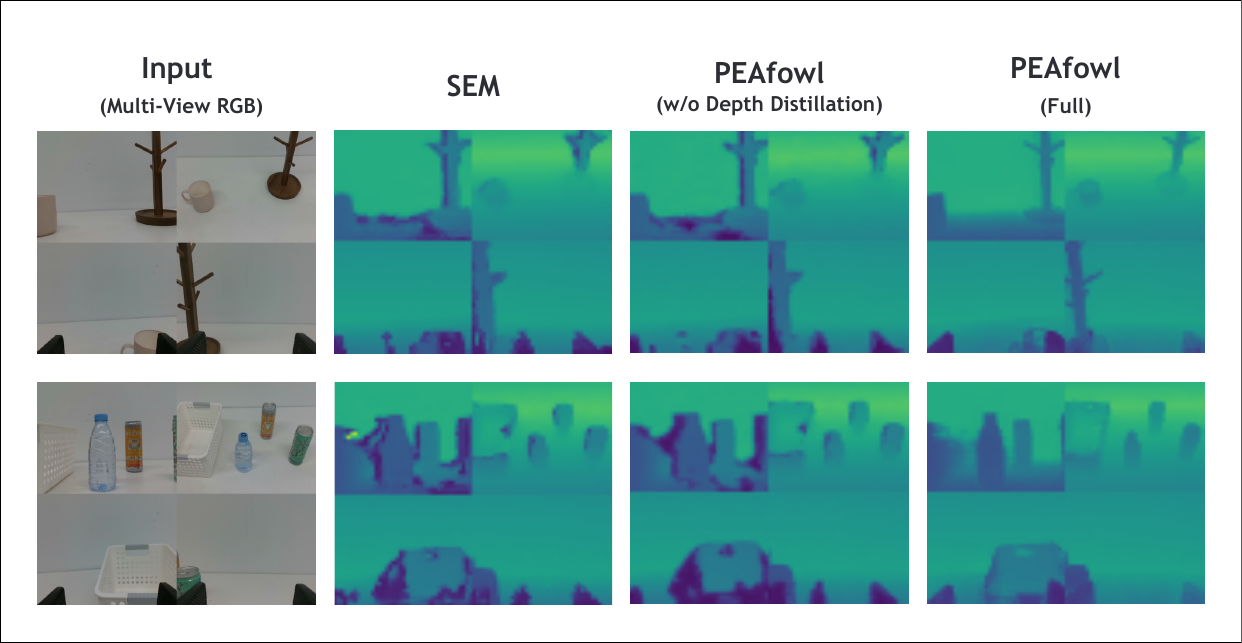}
  \caption{\textbf{Depth-distribution predictions (real robot).} 
  Predicted per-token depth distributions for SEM, PEAfowl w/o depth distillation, and full PEAfowl. PEAfowl produces cleaner predictions than SEM, and training-only depth distillation further sharpens and completes depth under noisy commodity sensors.}
  \label{fig:fig6}  
\end{figure}

\paragraph{Baselines}

We compare PEAfowl with visuomotor and bimanual VLA baselines.
Visuomotor policies include ACT~\cite{Zhao-RSS-23}, DP~\cite{Chi-RSS-23}, and DP3~\cite{Ze-RSS-24}, trained separately per task.
VLA baselines trained jointly across tasks include $\pi_0$~\cite{BlackK-RSS-25}, RDT~\cite{liu2025rdt}, and SEM~\cite{zhou2025sem}.
All baselines use the same observation setup, training splits, and sim-to-real fine-tuning protocol as PEAfowl.

\paragraph{Metrics}
In simulation, we run $100$ trials per task per setting, and on real-robot experiments $10$ trials per task. We report \textbf{Success Rate}, the fraction of trials that finish the task within the episode time limit.

\subsection{Results and Analysis}

\paragraph{Simulation Results}
As shown in Tab.~\ref{tab:sim_main_singlecol}, under the RoboTwin~2.0 DR setting, PEAfowl achieves an average success rate of \textbf{47.1\%} across $9$ tasks, consistently outperforming all baselines. In contrast, task-specific visuomotor policies trained per task rarely succeed on these challenging tasks, suggesting that 50 demonstrations are insufficient \rev{for these task-specific baselines} to handle appearance and layout variations.

\begin{table}[!t]
  \centering
  \scriptsize
  \setlength{\tabcolsep}{2.0pt}
  \renewcommand{\arraystretch}{1.08}
  \caption{\textbf{Simulation results on RoboTwin 2.0.} Success rate (\%) under Clean and DR (domain-randomized) settings. Task abbreviations: OM=\textbf{Open Microwave}, SB3=\textbf{Stack Blocks Three}, SW3=\textbf{Stack Bowls Three}, BR-RGB=\textbf{Blocks Ranking RGB}, BR-Size=\textbf{Blocks Ranking Size}, HM=\textbf{Hanging Mug}, OL=\textbf{Open Laptop}, PBF=\textbf{Place Burger Fries}, PEC=\textbf{Place Empty Cup}.}
  \label{tab:sim_main_singlecol}
  \begin{tabular}{>{\raggedright\arraybackslash}p{1.3cm}p{1.3cm}ccccccc}
    \toprule
    \textbf{Horizon} & \textbf{Task} & \textbf{DP} & \textbf{DP3} & \textbf{ACT} & \textbf{$\pi_0$} & \textbf{RDT} & \textbf{SEM} & \textbf{PEAfowl} \\
    \midrule
    \multicolumn{9}{c}{\textbf{Clean}} \\
    \midrule
    Long & OM      & 5  & 61 & \best{86} & 25 & 9  & 17 & \second{70} \\
         & SB3     & 0  & 1  & 0           & 2  & 0  & \second{29} & \best{72} \\
         & SW3     & 63 & 57 & 48          & 47 & 19 & \second{68} & \best{79} \\
         & BR-RGB  & 0  & 3  & 1           & 16 & 0  & \second{55} & \best{84} \\
         & BR-Size & 1  & 2  & 0           & 4  & 0  & \second{21} & \best{27} \\
         & Long Avg.& 13.8 & 24.8 & 27.0 & 18.8 & 5.6 & \second{38.0} & \best{66.4} \\
    \addlinespace
    Mid  & HM      & 8  & 17 & 7           & 6  & 0  & \second{24} & \best{41} \\
         & OL      & 49 & \best{82} & 56 & 38 & 25 & 75 & \second{77} \\
         & Mid Avg.& 28.5 & \second{49.5} & 31.5 & 22.0 & 12.5 & \second{49.5} & \best{59.0} \\
    \addlinespace
    Short& PBF     & 72 & 72 & 49          & 28 & 19 & \second{93} & \best{95} \\
         & PEC     & 37 & 65 & 61          & 32 & 23 & \second{77} & \best{81} \\
         & Short Avg.& 54.5 & 68.5 & 55.0 & 30.0 & 21.0 & \second{85.0} & \best{88.0} \\
         & Avg.    & 26.1 & 40.0 & 34.2 & 22.0 & 10.6 & \second{51.0} & \best{69.6} \\
    \midrule
    \multicolumn{9}{c}{\textbf{DR}} \\
    \midrule
    Long & OM      & 0  & 22 & 17 & 8  & 8  & \second{25} & \best{34} \\
         & SB3     & 0  & 0  & 0  & \second{15} & 0 & 1 & \best{34} \\
         & SW3     & 30 & 8  & 8  & \second{48} & 6 & 41 & \best{66} \\
         & BR-RGB  & 0  & 0  & 0  & 4  & 0  & \second{15} & \best{47} \\
         & BR-Size & 0  & 1  & 0  & \second{10} & 0 & 0 & \best{14} \\
         & Long Avg.& 6.0 & 6.2 & 5.0 & \second{17.0} & 2.8 & 16.4 & \best{39.0} \\
    \addlinespace
    Mid  & HM      & 1  & 9  & 2  & 6  & 3  & \second{18} & \best{26} \\
         & OL      & 32 & \second{61} & 18 & 45 & 31 & \second{61} & \best{68} \\
         & Mid Avg.& 16.5 & 35.0 & 10.0 & 25.5 & 17.0 & \second{39.5} & \best{47.0} \\
    \addlinespace
    Short& PBF     & 38 & 21 & 22 & 31 & 9  & \second{46} & \best{65} \\
         & PEC     & 1  & 5  & 3  & \second{32} & 3 & 10 & \best{70} \\
         & Short Avg.& 19.5 & 13.0 & 12.5 & \second{31.5} & 6.0 & 28.0 & \best{67.5} \\
         & Avg.    & 11.3 & 14.1 & 7.8 & 22.1 & 6.7 & \second{24.1} & \best{47.1} \\
    \bottomrule
  \end{tabular}
\end{table}

\begin{table}[t]
  \centering
  \scriptsize
  \setlength{\tabcolsep}{3.5pt}
  \renewcommand{\arraystretch}{1.08}
  \caption{\textbf{In-family held-out transfer results.}
  Success rates (\%) on two held-out RoboTwin task variants under the Clean and DR settings. 
  Models are trained on the corresponding nine-task suite and evaluated on SB2 and SW2. 
  Task abbr.: SB2=\textbf{Stack Blocks Two}, SW2=\textbf{Stack Bowls Two}.}
  \label{tab:in_domain_generalization}
  \begin{tabular}{>{\raggedright\arraybackslash}p{0.95cm}
                  >{\raggedright\arraybackslash}p{1.25cm}
                  cccc}
    \toprule
    \textbf{Task} & \textbf{Setting} & \textbf{$\pi_0$} & \textbf{RDT} & \textbf{SEM} & \textbf{PEAfowl} \\
    \midrule
    \multirow{2}{*}{SB2}
      & Clean & 25 & 0  & \second{49} & \best{74} \\
      & DR    & \second{34} & 0  & 14 & \best{51} \\
    \midrule
    \multirow{2}{*}{SW2}
      & Clean & 83 & 38 & \second{93} & \best{97} \\
      & DR    & \second{79} & 23 & 65 & \best{82} \\
    \bottomrule
  \end{tabular}
\end{table}

\begin{table}[t]
  \centering
  \scriptsize
  \setlength{\tabcolsep}{2.8pt}
  \renewcommand{\arraystretch}{1.08}
  \caption{\textbf{Real-world results.} Success rates (\%) on the dual-arm platform. Task abbr.: PS=\textbf{Place Shoe}, PBD=\textbf{Put Bottles Dustbin}.}
  \label{tab:real_world}
  \begin{tabular}{>{\raggedright\arraybackslash}p{0.95cm}ccc}
    \toprule
    \textbf{Task} & \textbf{PEAfowl} & \textbf{PEAfowl w/o DD} & \textbf{SEM} \\
    \midrule
    SW3 & \best{100 (10/10)} & \second{70 (7/10)} & 30 (3/10) \\
    HM  & \best{30 (3/10)}   & \second{10 (1/10)} & 0 (0/10) \\
    PBF & \best{60 (6/10)}   & \second{30 (3/10)} & 10 (1/10) \\
    PEC & \best{80 (8/10)}   & \second{50 (5/10)} & 20 (2/10) \\
    PS  & \best{60 (6/10)}   & \second{30 (3/10)} & 10 (1/10) \\
    PBD & \best{80 (8/10)}   & \second{30 (3/10)} & 0 (0/10) \\
    \midrule
    \textbf{Avg.} & \best{68.3} & \second{36.7} & 11.7 \\
    \bottomrule
  \end{tabular}

\end{table}

\begin{figure}[t]
  \centering
  \includegraphics[width=0.8\linewidth]{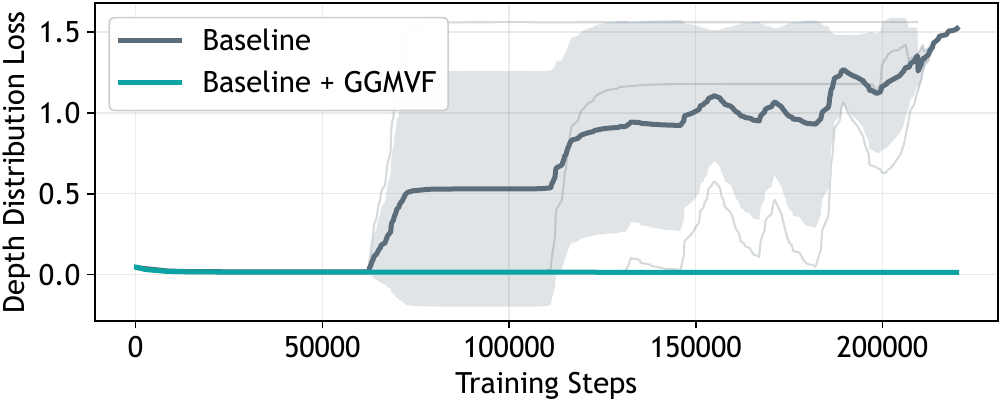}
  \caption{Depth-loss stability. 
  Under multi-task training, the Baseline’s depth loss diverges, while adding GGMVF stabilizes depth-distribution learning.}
  \label{fig:fig7}
\end{figure}

Among multi-task bimanual VLA baselines, $\pi_0$ achieves an average of $22.1\%$ under the DR setting, indicating that large-scale vision-language and robot-data pretraining improve \rev{cross-scene robustness}, whereas RDT underperforms under heavy randomization. SEM is the strongest baseline, which leverages 3D spatial position embeddings along with a joint-centric state encoder and action decoder.

PEAfowl improves over SEM by \textbf{23.0 pp}, with particularly large gains on long-horizon and occlusion-heavy tasks. Notably, on visually similar yet instruction-distinct tasks (\textbf{Stack Blocks Three} vs.\ \textbf{Blocks Ranking RGB}), $\pi_0$ and SEM exhibit task-specific biases, whereas PEAfowl performs well on both, suggesting \rev{more effective instruction-conditioned visual readout} and robustness to clutter and occlusions.

We further visualize multi-view token embeddings before and after cross-view aggregation using t-SNE as shown in Fig.~\ref{fig:fig5}. 
\rev{After aggregation, tokens associated with visually corresponding regions across views form more coherent clusters, qualitatively supporting that geometry-guided fusion promotes cross-view feature alignment under distractors and viewpoint changes.}

In the clean setting, which evaluates in-distribution task performance, PEAfowl achieves \textbf{69.6\%} average success, surpassing SEM by \textbf{18.6 pp}. Improvements are most obvious on long-horizon tasks requiring consistent spatial awareness under repeated occlusions and viewpoint changes, while gains on short-horizon tasks are smaller due to high baseline performance. Consistent with the DR evaluation, PEAfowl better identifies instruction-relevant targets and relations in referentially ambiguous scenes.

Finally, we evaluate \rev{in-family held-out transfer} on two additional tasks not included in the nine-task training set in Tab.~\ref{tab:in_domain_generalization}. PEAfowl consistently outperforms all baselines under both clean and DR settings. Notably, for \textbf{Stack Blocks Two} under DR, SEM drops from $49\%$ to $14\%$, while PEAfowl maintains $51\%$, \rev{suggesting better transfer to unseen but related bimanual tasks.}

\paragraph{Real-World Results}

Tab.~\ref{tab:real_world} compares PEAfowl with SEM, the strongest baseline in simulation, and further ablates training-only depth distillation (DD). PEAfowl outperforms SEM on all $6$ real-robot tasks, \rev{indicating improved real-robot performance under commodity RGB-D sensing.} DD brings additional improvements, particularly on depth-sensitive tasks such as \textbf{Put Bottles Dustbin} and \textbf{Hanging Mug}. Fig.~\ref{fig:fig6} visualizes an example of depth-distribution prediction, \rev{suggesting that distillation injects geometric priors from a pretrained depth teacher into the depth-distribution head}.

We also observe that SEM suffers depth-loss divergence during real-robot fine-tuning (cf.\ Section~\ref{ablations}), highlighting the difficulty of learning stable depth-aware representations from noisy sensor depth alone. In contrast, PEAfowl’s local RGB-D pairwise fusion and cross-view 3D neighbor aggregation mitigate unreliable depth and stabilize \rev{multi-view spatial representation learning}. Qualitative rollouts are provided in the supplementary video.

\begin{table}[!t]
  \centering
  \scriptsize
  \setlength{\tabcolsep}{2.4pt}
  \renewcommand{\arraystretch}{1.08}
  \caption{\textbf{Ablations on RoboTwin 2.0.} Success rate (\%) on nine tasks under Clean and DR. Task abbreviations follow Table~\ref{tab:sim_main_singlecol}.}
  \label{tab:ablation}
  \begin{tabular}{>{\raggedright\arraybackslash}p{1.05cm}ccccc}
    \toprule
    \textbf{Task} &
    \textbf{PEAfowl} &
    \textbf{\shortstack{w/1-step\\ SimAttn}} &
    \textbf{\shortstack{w/o\\ VG-Text}} &
    \textbf{\shortstack{Baseline\\ + GGMVF}} &
    \textbf{Baseline} \\
    \midrule
    \multicolumn{6}{c}{\textbf{Clean}} \\
    \midrule
    OM      & \best{70} & 44 & \second{65} & 45 & 17 \\
    SB3     & \best{72} & \second{70} & 39 & 21 & 1 \\
    SW3     & \best{79} & 71 & 75 & \second{77} & 50 \\
    BR-RGB  & \best{84} & \second{75} & 62 & 69 & 44 \\
    BR-Size & \best{27} & \second{26} & \best{27} & 21 & 16 \\
    HM      & \best{41} & 38 & \second{40} & 39 & 29 \\
    OL      & \best{77} & 74 & \second{76} & \second{76} & \second{76} \\
    PBF     & \second{95} & \best{98} & \best{98} & 94 & \second{95} \\
    PEC     & \best{81} & \second{79} & \second{79} & 77 & 74 \\
    \addlinespace
    \textbf{Avg.} & \best{69.6} & \second{63.9} & 62.3 & 57.7 & 44.7 \\
    \midrule
    \multicolumn{6}{c}{\textbf{DR}} \\
    \midrule
    OM      & \best{34} & 31 & \second{32} & 26 & 17 \\
    SB3     & \best{34} & 18 & \second{26} & 7 & 1 \\
    SW3     & \best{66} & \second{61} & \second{61} & \second{61} & 50 \\
    BR-RGB  & \best{47} & \second{35} & 31 & 34 & 11 \\
    BR-Size & \best{14} & 1 & 3 & \second{7} & 0 \\
    HM      & \best{26} & 19 & 20 & \second{23} & 14 \\
    OL      & \best{68} & \second{65} & \best{68} & \best{68} & \best{68} \\
    PBF     & \best{65} & \second{49} & \second{49} & 44 & \best{65} \\
    PEC     & \best{70} & 46 & \second{60} & 52 & 56 \\
    \addlinespace
    \textbf{Avg.} & \best{47.1} & 36.1 & \second{38.9} & 35.8 & 31.3 \\
    \bottomrule
  \end{tabular}
\end{table}

\subsection{Ablation Studies}
\label{ablations}

We ablate PEAfowl on RoboTwin~2.0 by removing individual components from the full model and by adding our geometry-guided perception to a simple visuomotor baseline. 
\rev{The baseline uses the same RGB-D encoders and SEM-style diffusion decoder as PEAfowl, but excludes both the language-guided readout and the geometry-guided operations, including pairwise RGB-D fusion and cross-view 3D neighbor aggregation.} We evaluate three variants: (i) replacing the Perceiver-style iterative readout with OTTER’s one-step similarity readout (\textbf{w/1-step SimAttn}); (ii) removing vision-grounded text tokens and conditioning the policy only on instruction-summary tokens (\textbf{w/o VG-Text}); and (iii) adding the full geometry-guided module to the baseline (\textbf{Baseline+GGMVF}).

As shown in Tab.~\ref{tab:ablation}, \textbf{w/1-step SimAttn} yields a small degradation under the clean setting but a substantially larger drop under DR, \rev{suggesting the effectiveness of iterative, text-aware evidence accumulation} under clutter and scene shifts. \textbf{w/o VG-Text} also lowers success, with more obvious drops on attribute- and reference-sensitive tasks, \rev{supporting the contribution of text-aware visual extraction}. Finally, the baseline suffers from depth-loss divergence and collapsed depth distributions. This instability is consistently observed across independent training runs with different random seeds under the same settings (Fig.~\ref{fig:fig7}). \textbf{Baseline+GGMVF} remains stable and improves long-horizon performance, \rev{indicating that our geometry-guided pathway contributes to more stable depth-aware multi-task learning}.

We further ablate camera views under the DR setting. As shown in Tab.~\ref{tab:mv_ablation_dr}, removing the front view reduces the average success rate from $47.1\%$ to $35.0\%$, and using only the head view further reduces it to $26.8\%$, confirming the importance of complementary front and wrist observations for bimanual manipulation.

\begin{table}[t]
  \centering
  \scriptsize
  \setlength{\tabcolsep}{3.0pt}
  \renewcommand{\arraystretch}{1.08}
  \caption{\revtext{\textbf{Average multi-view ablation under the DR setting.}}
  Success rate (\%). ``w/o Front'' removes the front view; ``Head-only'' removes the front and both wrist views.}
  \label{tab:mv_ablation_dr}
  \begin{tabular}{lccc}
    \toprule
    \textbf{Setting} & \textbf{PEAfowl} & \textbf{w/o Front} & \textbf{Head-only} \\
    \midrule
    \textbf{Avg.} & \revtext{47.1} & \revtext{35.0} & \revtext{26.8} \\
    \bottomrule
  \end{tabular}
\end{table}

\section{CONCLUSIONS}

In this paper, we proposed PEAfowl, a perception-enhanced multi-view VLA policy for bimanual manipulation that \rev{improves robustness} under clutter, occlusions, and scene variations in the evaluated settings. PEAfowl couples geometry-guided fusion for depth-aware cross-view feature aggregation with a Perceiver-style text-as-query readout over frozen CLIP features for improved \rev{instruction-conditioned visual extraction}. Across RoboTwin~2.0 and real-robot evaluations, \rev{PEAfowl improves over the evaluated bimanual VLA and visuomotor baselines, with ablations highlighting the contributions of geometry-guided fusion, language-guided readout, depth distillation, and multi-view sensing}.

\bibliographystyle{IEEEtran}
\bibliography{mybibfile}

@inproceedings{brohan2023rt2,
  author    = {Brianna Zitkovich and Tianhe Yu and Sichun Xu and Peng Xu and Ted Xiao and Fei Xia and Jialin Wu and Paul Wohlhart and Stefan Welker and Ayzaan Wahid and others},
  title     = {{RT-2}: Vision-Language-Action Models Transfer Web Knowledge to Robotic Control},
  booktitle = {Proceedings of The 7th CoRL},
  volume    = {229},
  pages     = {2165--2183},
  year      = {2023},
  publisher = {PMLR},
}

@inproceedings{pmlr-v270-kim25c,
  author    = {Kim, Moo Jin and Pertsch, Karl and Karamcheti, Siddharth and Xiao, Ted and Balakrishna, Ashwin and Nair, Suraj and Rafailov, Rafael and Foster, Ethan P and Sanketi, Pannag R and Vuong, Quan and Kollar, Thomas and Burchfiel, Benjamin and Tedrake, Russ and Sadigh, Dorsa and Levine, Sergey and Liang, Percy and Finn, Chelsea},
  title     = {OpenVLA: An Open-Source Vision-Language-Action Model},
  booktitle = {Proceedings of The 8th CoRL},
  volume    = {270},
  pages     = {2679--2713},
  year      = {2025},
  publisher = {PMLR},
}

@INPROCEEDINGS{BlackK-RSS-25, 
    AUTHOR    = {Kevin Black AND Noah Brown AND Danny Driess AND Adnan Esmail AND Michael Robert Equi AND Chelsea Finn AND Niccolo Fusai AND Lachy Groom AND Karol Hausman AND Brian Ichter AND others}, 
    TITLE     = {{$\pi_0$: A Vision-Language-Action Flow Model for General Robot Control}}, 
    BOOKTITLE = {Proceedings of RSS}, 
    YEAR      = {2025}, 
    ADDRESS   = {LosAngeles, CA, USA}, 
    MONTH     = {June}, 
    DOI       = {10.15607/RSS.2025.XXI.010} 
}

@INPROCEEDINGS{QuD-RSS-25, 
    AUTHOR    = {Delin Qu AND Haoming Song AND Qizhi Chen AND Yuanqi Yao AND Xinyi Ye AND Jiayuan Gu AND Zhigang Wang AND Yan Ding AND Bin Zhao AND Dong Wang AND Xuelong Li}, 
    TITLE     = {{SpatialVLA: Exploring Spatial Representations for Visual-Language-Action Models}}, 
    BOOKTITLE = {Proceedings of RSS}, 
    YEAR      = {2025}, 
    ADDRESS   = {LosAngeles, CA, USA}, 
    MONTH     = {June}, 
    DOI       = {10.15607/RSS.2025.XXI.011} 
}

@inproceedings{liu2025rdt,
  author    = {Songming Liu and Lingxuan Wu and Bangguo Li and Hengkai Tan and Huayu Chen and Zhengyi Wang and Ke Xu and Hang Su and Jun Zhu},
  title     = {{RDT-1B}: A Diffusion Foundation Model for Bimanual Manipulation},
  booktitle = {ICLR},
  year      = {2025},
}

@INPROCEEDINGS{Zhao-RSS-23, 
    AUTHOR    = {Tony Z. Zhao AND Vikash Kumar AND Sergey Levine AND Chelsea Finn}, 
    TITLE     = {{Learning Fine-Grained Bimanual Manipulation with Low-Cost Hardware}}, 
    BOOKTITLE = {Proceedings of RSS}, 
    YEAR      = {2023}, 
    ADDRESS   = {Daegu, Republic of Korea}, 
    MONTH     = {July}, 
    DOI       = {10.15607/RSS.2023.XIX.016} 
}

@INPROCEEDINGS{Chi-RSS-23, 
    AUTHOR    = {Cheng Chi AND Siyuan Feng AND Yilun Du AND Zhenjia Xu AND Eric Cousineau AND Benjamin CM Burchfiel AND Shuran Song}, 
    TITLE     = {{Diffusion Policy: Visuomotor Policy Learning via Action Diffusion}}, 
    BOOKTITLE = {Proceedings of RSS}, 
    YEAR      = {2023}, 
    ADDRESS   = {Daegu, Republic of Korea}, 
    MONTH     = {July}, 
    DOI       = {10.15607/RSS.2023.XIX.026} 
}

@INPROCEEDINGS{Ze-RSS-24, 
    AUTHOR    = {Yanjie Ze AND Gu Zhang AND Kangning Zhang AND Chenyuan Hu AND Muhan Wang AND Huazhe Xu}, 
    TITLE     = {{3D Diffusion Policy: Generalizable Visuomotor Policy Learning via Simple 3D Representations}}, 
    BOOKTITLE = {Proceedings of RSS}, 
    YEAR      = {2024}, 
    ADDRESS   = {Delft, Netherlands}, 
    MONTH     = {July}, 
    DOI       = {10.15607/RSS.2024.XX.067} 
}

@inproceedings{seita2023mvmwm,
  author    = {Younggyo Seo and Junsu Kim and Stephen James and Kimin Lee and Jinwoo Shin and Pieter Abbeel},
  title     = {Multi-View Masked World Models for Visual Robotic Manipulation},
  booktitle = {Proceedings of the 40th ICML},
  volume    = {202},
  pages     = {30613--30632},
  year      = {2023},
  publisher = {PMLR},
}

@INPROCEEDINGS{Goyal-RSS-24, 
    AUTHOR    = {Ankit Goyal AND Valts Blukis AND Jie Xu AND Yijie Guo AND Yu-Wei Chao AND Dieter Fox}, 
    TITLE     = {{RVT-2: Learning Precise Manipulation from Few Demonstrations}}, 
    BOOKTITLE = {Proceedings of RSS}, 
    YEAR      = {2024}, 
    ADDRESS   = {Delft, Netherlands}, 
    MONTH     = {July}, 
    DOI       = {10.15607/RSS.2024.XX.055} 
}

@inproceedings{li2025threedmvp,
  author    = {Qian, Shengyi and Mo, Kaichun and Blukis, Valts and Fouhey, David F. and Fox, Dieter and Goyal, Ankit},
  title     = {3D-MVP: 3D Multiview Pretraining for Manipulation},
  booktitle = {Proceedings of CVPR},
  year      = {2025},
  pages     = {22530--22539},
}

@inproceedings{liu2024groundingdino,
  author    = {Shilong Liu and Zhaoyang Zeng and Tianhe Ren and Feng Li and Hao Zhang and Jie Yang and Qing Jiang and Chunyuan Li and Jianwei Yang and Hang Su and others},
  title     = {Grounding {DINO}: Marrying {DINO} with Grounded Pre-Training for Open-Set Object Detection},
  booktitle = {Proceedings of ECCV},
  year      = {2024},
  doi       = {10.1007/978-3-031-72970-6_3}
}

@inproceedings{zhu2024clearclip,
  author    = {Mengcheng Lan and Chaofeng Chen and Yiping Ke and Xinjiang Wang and Litong Feng and Wayne Zhang},
  title     = {Clear{CLIP}: Decomposing {CLIP} Representations for Dense Vision-Language Inference},
  booktitle = {Proceedings of ECCV},
  year      = {2024}
}

@inproceedings{radford2021clip,
  author    = {Alec Radford and Jong Wook Kim and Chris Hallacy and Aditya Ramesh and Gabriel Goh and Sandhini Agarwal and Girish Sastry and Amanda Askell and Pamela Mishkin and Jack Clark and others},
  title     = {Learning Transferable Visual Models From Natural Language Supervision},
  booktitle = {Proceedings of the 38th ICML},
  volume    = {139},
  year      = {2021},
  publisher = {PMLR}
}

@inproceedings{huang2025otter,
  author    = {Huang Huang and Fangchen Liu and Letian Fu and Tingfan Wu and Mustafa Mukadam and Jitendra Malik and Ken Goldberg and Pieter Abbeel},
  title     = {{OTTER}: A Vision-Language-Action Model with Text-Aware Visual Feature Extraction},
  booktitle = {Proceedings of the 42nd ICML},
  volume    = {267},
  year      = {2025},
  publisher = {PMLR},
}

@inproceedings{jaegle2021perceiverio,
  author = {Andrew Jaegle and Sebastian Borgeaud and Jean-Baptiste Alayrac and Carl Doersch and Catalin Ionescu and David Ding and Skanda Koppula and Daniel Zoran and Andrew Brock and Evan Shelhamer and Olivier J. Henaff and Matthew Botvinick and Andrew Zisserman and Oriol Vinyals and Joao Carreira},
  title     = {Perceiver {IO}: A General Architecture for Structured Inputs \& Outputs},
  booktitle = {ICLR},
  year      = {2022},
}

@inproceedings{bachlechner2021rezero,
  author    = {Thomas Bachlechner and Bodhisattwa Prasad Majumder and Henry Mao and Gary Cottrell and Julian McAuley},
  title     = {{ReZero} is All You Need: Fast Convergence at Large Depth},
  booktitle = {Proceedings of the 37th UAI},
  year      = {2021},
  volume    = {161},
  publisher = {PMLR},
}

@inproceedings{he2016resnet,
  author    = {Kaiming He and Xiangyu Zhang and Shaoqing Ren and Jian Sun},
  title     = {Deep Residual Learning for Image Recognition},
  booktitle = {Proceedings of CVPR},
  year      = {2016},
  pages     = {770--778},
  doi = {10.1109/CVPR.2016.90}
}

@inproceedings{chen2025robotwin,
  title     = {{RoboTwin} 2.0: A Scalable Data Generator and Benchmark with Strong Domain Randomization for Robust Bimanual Robotic Manipulation},
  author    = {Tianxing Chen and Zanxin Chen and Baijun Chen and Zijian Cai
               and Yibin Liu and Zixuan Li and Qiwei Liang and Xianliang Lin
               and Yiheng Ge and Zhenyu Gu and Weiliang Deng and Yubin Guo
               and Tian Nian and Xuanbing Xie and Qiangyu Chen and Kailun Su
               and Tianling Xu and Guodong Liu and Mengkang Hu and Huan-ang Gao
               and Kaixuan Wang and Zhixuan Liang and Yusen Qin and Xiaokang Yang
               and Ping Luo and Yao Mu},
  booktitle = {Proceedings of the 43rd ICML},
  year      = {2026},
}

@misc{zhou2025sem,
  author       = {Xuewu Lin and Tianwei Lin and Lichao Huang and Hongyu Xie and Yiwei Jin and Keyu Li and Zhizhong Su},
  title        = {{SEM}: Enhancing Spatial Understanding for Robust Robot Manipulation},
  howpublished = {arXiv preprint arXiv:2505.16196},
  year         = {2025},
}

@inproceedings{liu2025manipulation,
 author = {Liu, Minghuan and Zhu, Zhengbang and Han, Xiaoshen and Hu, Peng and Lin, Haotong and Li, Xinyao and Chen, Jingxiao and Xu, Jiafeng and Yang, Yichu and Lin, Yunfeng and Li, Xinghang and Yu, Yong  and Zhang, Weinan and Kong, Tao and Kang, Bingyi},
 booktitle = {ICLR},
 pages = {11238--11271},
 title = {Manipulation as in Simulation: Enabling Accurate Geometry Perception in Robots},
 volume = {2026},
 year = {2026}
}

@InProceedings{wang2025vggt,
  author    = {Wang, Jianyuan and Chen, Minghao and Karaev, Nikita and Vedaldi, Andrea and Rupprecht, Christian and Novotny, David},
  title     = {VGGT: Visual Geometry Grounded Transformer},
  booktitle = {Proceedings of CVPR},
  month     = {June},
  year      = {2025},
  pages     = {5294-5306}
}

@INPROCEEDINGS{fan2025neugrasp,
  author={Fan, Qingyu and Cai, Yinghao and Li, Chao and He, Wenzhe and Zheng, Xudong and Lu, Tao and Liang, Bin and Wang, Shuo},
  booktitle={IEEE ICRA}, 
  title={NeuGrasp: Generalizable Neural Surface Reconstruction with Background Priors for Material-Agnostic Object Grasp Detection}, 
  year={2025},
  volume={},
  number={},
  pages={3197-3203},
  doi={10.1109/ICRA55743.2025.11127348}
}

\clearpage
\appendices

\section{Additional Implementation Details}
\label{app:implementation}

\subsection{Core Model and Training Settings}
\label{app:core_settings}

Table~\ref{tab:implementation_details} summarizes the core implementation settings. PEAfowl uses a single-step observation history and predicts an action chunk with horizon $H=64$. During deployment, the policy is queried in a receding-horizon manner. The geometry-guided branch predicts per-token depth distributions over $B=128$ linearly spaced depth bins in $[0.01,1.2]$\,m, corresponding to a bin interval of approximately $9.4$\,mm. Cross-view 3D aggregation uses up to $K=16$ nearest tokens from other camera views (or all available candidates when fewer than 16 exist) and a distance-softmax temperature $\tau=0.08$\,m.

\begin{table}[t]
\centering
\caption{\textbf{Implementation details of PEAfowl.}}
\label{tab:implementation_details}
\scriptsize
\setlength{\tabcolsep}{3pt}
\renewcommand{\arraystretch}{1.08}
\begin{tabularx}{\linewidth}{@{}p{0.22\linewidth}p{0.28\linewidth}X@{}}
\toprule
\textbf{Category} & \textbf{Item} & \textbf{Setting} \\
\midrule
Observation & History length & $H_{\mathrm{obs}}=1$ \\
Observation & RGB-D input resolution & $320\times256$ \\
Observation & Number of camera views & $4$ \\
Policy output & Prediction horizon & $H=64$ \\
Policy output & Valid action chunk & $32$ steps \\
Token dimension & Hidden dimension & $d=256$ \\
RGB encoder & Backbone & Swin-T initialized from Grounding-DINO \\
Depth encoder & Backbone & ResNet-34 \\
Geometry branch & Feature strides & $(8,16,32,64)$ \\
Geometry branch & Decoder-attended levels & stride-16 and stride-32 \\
Depth distribution & Depth bins & $B=128$ \\
Depth distribution & Depth range & $[0.01,1.2]$\,m \\
Depth distribution & Bin interval & $\approx 9.4$\,mm \\
Depth supervision & Token-validity threshold & $0.5$ \\
Cross-view aggregation & Neighbor cap & $K=16$ \\
Cross-view aggregation & Distance temperature & $\tau=0.08$\,m \\
Cross-view aggregation & Residual gate initialization & $0.5$ \\
RGB-D pair fusion & Attention block & 1 block, 4 heads \\
RGB-D pair fusion & Fusion dimension & $d_{\mathrm{dep}}=32$ \\
RGB-D pair fusion & FFN hidden dimension & 128 \\
Language branch & CLIP backbone & frozen CLIP ViT-L/14 \\
Language branch & Retained text tokens & $K_{\mathrm{txt}}=64$ \\
Language branch & Perceiver-style latent blocks & $M=3$ per view \\
Language branch & Readout tokens & $R=64$ per view \\
Language branch & ReZero gate initialization & $0$ \\
Action decoder & Diffusion decoder blocks & 6 \\
Diffusion training & Scheduler & DDPM \\
Diffusion training & Timesteps & 1000 \\
Diffusion training & Beta schedule & squared cosine \\
Diffusion training & Prediction type & sample prediction \\
Diffusion inference & Scheduler & DPMSolverMultistep \\
Diffusion inference & Denoising steps & 10 \\
Main simulation optimization & Optimizer & AdamW \\
Main simulation optimization & Batch size & 16 \\
Main simulation optimization & Training updates & $4\times10^5$ \\
Main simulation optimization & Base learning rate & $1.41\times10^{-4}$ \\
Main simulation optimization & Weight decay & $5\times10^{-4}$ \\
Main simulation optimization & RGB-backbone LR multiplier & $0.1\times$ \\
Main simulation optimization & Warmup & 500 iterations \\
Main simulation optimization & LR decay & $\times 0.1$ at 90\% training \\
\bottomrule
\end{tabularx}
\end{table}

\begin{table*}[t]
\centering
\caption{
\textbf{Robustness results under the separately trained robustness-study configuration.}
Success rate (\%) on RoboTwin 2.0. SEM is reproduced under the same training setting as a matched no-perturbation reference. PEAfowl is further evaluated under calibration and depth perturbations without retraining. Each result is averaged over nine tasks with 100 rollouts per task. Parentheses indicate the change from the unperturbed PEAfowl checkpoint under the same split.
}
\label{tab:supp_geometry_robustness}
\scriptsize
\setlength{\tabcolsep}{4.5pt}
\renewcommand{\arraystretch}{1.08}
\begin{tabular}{llccc}
\toprule
\textbf{Method} &
\textbf{Perturbation} &
\textbf{Level} &
\textbf{Clean SR} &
\textbf{DR SR} \\
\midrule
SEM
    & None
    & --     
    & 46.3
    & 11.8 \\
\midrule
PEAfowl
    & None
    & --     
    & \textbf{72.4}
    & \textbf{38.3} \\
\midrule
PEAfowl
    & Calibration
    & Mild   
    & 66.4 ($-6.0$)
    & 36.2 ($-2.1$)  \\
PEAfowl
    & Calibration
    & Medium 
    & 49.2 ($-23.2$)
    & 31.9 ($-6.4$)  \\
PEAfowl
    & Calibration
    & Severe 
    & 32.4 ($-40.0$)
    & 25.0 ($-13.3$) \\
\midrule
PEAfowl
    & Gaussian depth noise
    & 5\,mm  
    & 74.7 ($+2.3$)
    & 38.2 ($-0.1$)  \\
PEAfowl
    & Gaussian depth noise
    & 10\,mm 
    & 71.2 ($-1.2$)
    & 38.0 ($-0.3$)  \\
PEAfowl
    & Gaussian depth noise
    & 20\,mm 
    & 72.4 ($0.0$)
    & 37.8 ($-0.5$)  \\
\midrule
PEAfowl
    & Missing depth
    & 10\%   
    & 43.2 ($-29.2$)
    & 24.8 ($-13.5$) \\
PEAfowl
    & Missing depth
    & 30\%   
    & 10.8 ($-61.6$)
    & 20.9 ($-17.4$) \\
PEAfowl
    & Missing depth
    & 50\%   
    & 7.2 ($-65.2$)
    & 7.1 ($-31.2$)  \\
\bottomrule
\end{tabular}
\end{table*}

\subsection{Pairwise RGB-D Token Fusion}
\label{app:pairwise_rgbd}

The pairwise RGB-D token fusion module implements the local fusion in Eq.~\ref{eq:pairwise_rgbd}. For each view $v$ and token index $n$, the co-located RGB and depth tokens are projected to the common fusion dimension $d_{\mathrm{dep}}=32$ and processed as a length-2 sequence. The module consists of one four-head self-attention layer followed by a feed-forward network with hidden dimension 128. We apply this local fusion only in the depth-distribution branch; the main RGB stream remains unchanged before cross-view aggregation.

\subsection{Depth Distillation Targets}
\label{app:depth_distillation_targets}

For real-robot fine-tuning, the pretrained Camera Depth Model (CDM) is used only as an offline teacher. Given the raw RGB-D observation $(\mathbf{I}^{(v)},\mathbf{D}^{(v)})$, CDM produces a refined depth map $\tilde{\mathbf{D}}^{(v)}$. The policy input remains the raw depth map $\mathbf{D}^{(v)}$ during both training and inference; the refined CDM output is used only to construct supervision for the predicted per-token depth distribution.

For each feature level, the refined pixel depths are first converted to soft two-hot depth-bin targets and then average-pooled according to the corresponding token stride, as described in Sec.~\ref{sec:depth_distill}. The valid-pixel fraction in each pooled region defines a token weight $\omega^{(v)}_n\in[0,1]$: fractions below the validity threshold of $0.5$ are set to zero, while retained tokens keep their continuous valid-pixel fraction. The predicted distributions are supervised by the validity-weighted soft-label BCE loss in Eq.~\ref{eq:loss_depth}.

The depth-distillation coefficient is $\lambda_{\mathrm{depth}}=1.0$. The diffusion and forward-kinematics regression terms use the same configured state-wise regression weights; the complete objective is given in Eq.~\ref{eq:loss_total}.

\subsection{Qualitative CDM Depth Refinement}
\label{app:cdm_depth_quality}

\begin{figure}[!t]
    \centering
    \subfloat[Put Bottles Dustbin from our real-robot data.\label{fig:raw_depth_bottle}]{%
        \includegraphics[width=0.4\linewidth]{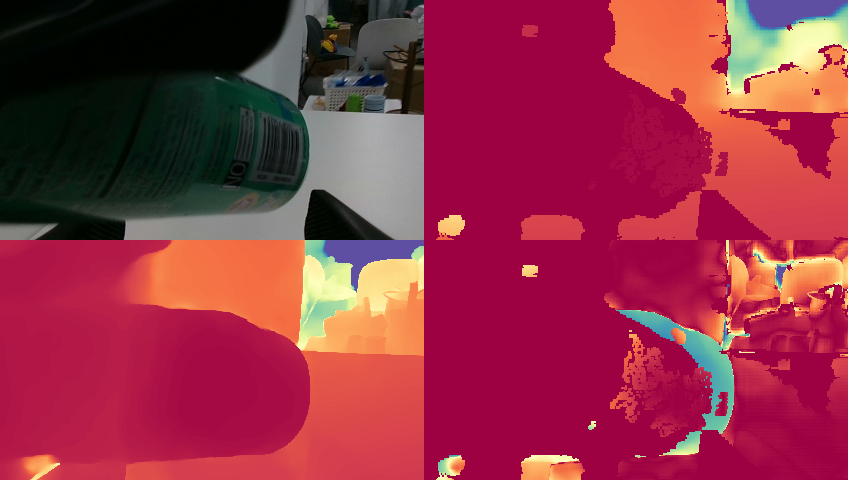}}
    \hfill
    \subfloat[Hanging Mug from our real-robot data.\label{fig:raw_depth_cup}]{%
        \includegraphics[width=0.4\linewidth]{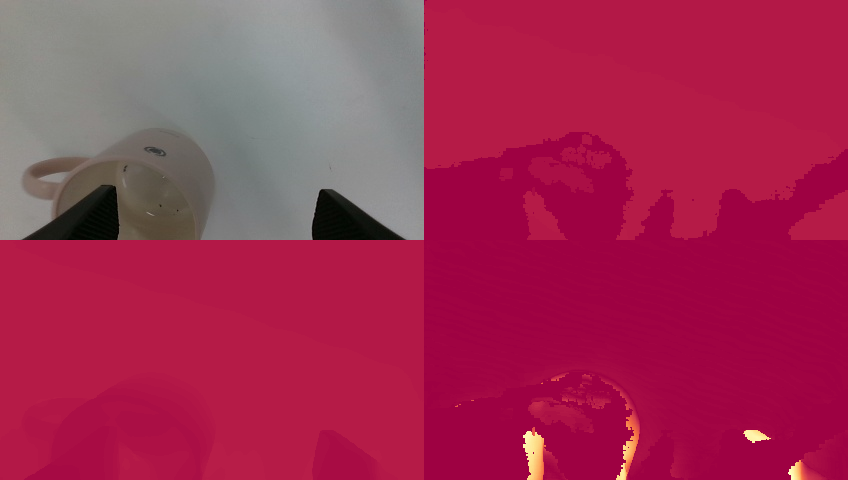}}
    \par\vspace{0.6em}
    \subfloat[Separately collected cluttered multi-object scene.\label{fig:raw_depth_clutter}]{%
        \includegraphics[width=0.6\linewidth]{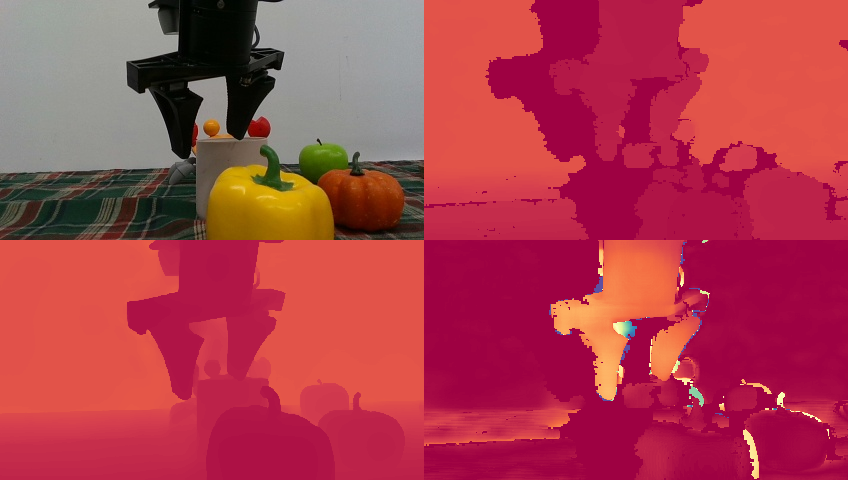}}
    \caption{
    \textbf{Qualitative comparison between raw Intel RealSense D435 depth and CDM-refined depth.}
    In each panel, the top row shows RGB and raw depth; the bottom row shows CDM-refined depth and the pixel-wise difference map. (a) and (b) are from real-robot data. (c) is a separately collected cluttered scene for visualization only and is not used in experimental results.
    }
    \label{fig:raw_d435_depth_quality}
\end{figure}

Fig.~\ref{fig:raw_d435_depth_quality} qualitatively compares raw Intel RealSense D435 depth maps with the corresponding CDM-refined maps. The examples illustrate how the offline teacher modifies missing or noisy regions used to construct token-level depth targets. CDM is used only to generate training supervision for the depth-distribution head and is not executed during policy inference.

\section{Additional Deployment Details}
\label{app:runtime}

The profiling setup in Tab.~\ref{tab:runtime_analysis} uses a single NVIDIA RTX 4090 GPU, batch size one, four $320\times256$ RGB-D views, 10 warm-up iterations, 100 timed iterations, and 10 diffusion denoising steps. Under this setup, SEM and PEAfowl have mean/p90 total latencies of 314.5/319.0\,ms and 325.1/332.0\,ms, respectively. PEAfowl therefore adds 10.6\,ms of mean latency (3.4\%) and increases peak allocated memory from 0.95\,GB to 3.04\,GB. CDM is excluded because it is used only for training-target generation. During real-robot deployment, the low-level controller runs at 25\,Hz and the high-level policy is queried at 1\,Hz, so the measured policy call remains within the 1\,s query interval.

\section{Additional Robustness Analysis}
\label{app:robustness}

\subsection{Training Configuration for Robustness Experiments}
\label{app:robustness_training}

The PEAfowl robustness experiments use separately trained Clean and DR checkpoints rather than the main-result checkpoints.
For each split, the corresponding PEAfowl robustness checkpoint is trained on two 48\,GB GPUs with a per-GPU batch size of 16, a global batch size of 32, and a learning rate of $2\times10^{-4}$. The Clean configuration uses a maximum of 152,712 updates over 28 epochs, whereas the DR configuration uses a maximum of 148,980 updates over 26 epochs.
Since this configuration differs from the main-result training setting, we use the results in this section only for within-checkpoint sensitivity analysis: each perturbed result is compared with the no-perturbation result from the corresponding split-specific PEAfowl checkpoint.

Each PEAfowl robustness checkpoint is trained on the same nine RoboTwin tasks with 50 demonstrations per task. 
The policy uses the same four RGB-D views, proprioceptive state, and 64-step action chunk as in the main experiments. 
Each evaluation setting is tested in closed-loop rollouts over the nine tasks, with 100 rollouts per task.

As matched no-perturbation references, we also reproduce SEM under the same task set, demonstration budget, observation setup, two-GPU training setup, batch size, and learning rate.
The reproduced SEM results are 46.3\% on Clean and 11.8\% on DR, respectively. 
Thus, the key comparisons in this section are PEAfowl versus SEM under the same no-perturbation robustness-study setting, and perturbed PEAfowl versus the no-perturbation PEAfowl result from the corresponding split-specific checkpoint.

\subsection{Perturbation Protocol}
\label{app:robustness_protocol}

After training, we evaluate each policy under controlled perturbations to camera calibration and depth observations. During evaluation, the environment dynamics, RGB observations, task initialization distribution, and policy weights remain fixed for a given checkpoint. Only the camera parameters or depth maps used by the policy are perturbed.

For calibration perturbation, each camera independently receives one fixed perturbation at the beginning of an episode, and the perturbation remains constant throughout the episode. The translation, rotation, focal-length, and principal-point perturbation variables are independently sampled from zero-mean Gaussian distributions. This models persistent calibration bias rather than frame-wise camera jitter:
\begin{equation}
\mathbf{T}'_{\mathrm{world}\rightarrow\mathrm{cam}}
=
\Delta\mathbf{T}\,
\mathbf{T}_{\mathrm{world}\rightarrow\mathrm{cam}},
\end{equation}
\begin{equation}
f_x'=f_x(1+\epsilon_{f_x}),\quad
f_y'=f_y(1+\epsilon_{f_y}).
\end{equation}
\begin{equation}
c_x'=c_x+\epsilon_{c_x},\quad
c_y'=c_y+\epsilon_{c_y}.
\end{equation}
The mild, medium, and severe settings respectively use translation standard deviations of $(5,15,30)$\,mm, rotation standard deviations of $(1,3,5)^\circ$, focal-length standard deviations of $(0.5,1.5,3.0)\%$, and principal-point standard deviations of $(2,5,10)$ pixels.

For additive depth noise, zero-mean Gaussian noise is added only to valid depth pixels:
\begin{equation}
d'=\max(d+\epsilon_d,0),\qquad
\epsilon_d\sim\mathcal{N}(0,\sigma_d^2),
\end{equation}
where $\sigma_d\in\{5,10,20\}$\,mm. Existing invalid pixels remain invalid.

For missing-depth corruption, we randomly set $10\%$, $30\%$, or $50\%$ of originally valid depth pixels to zero while keeping the RGB observations and camera parameters unchanged. The evaluation script uses a fixed default perturbation seed of 2026 for these controlled perturbations.

\begin{figure}[t]
    \centering
    \includegraphics[width=0.85\linewidth]{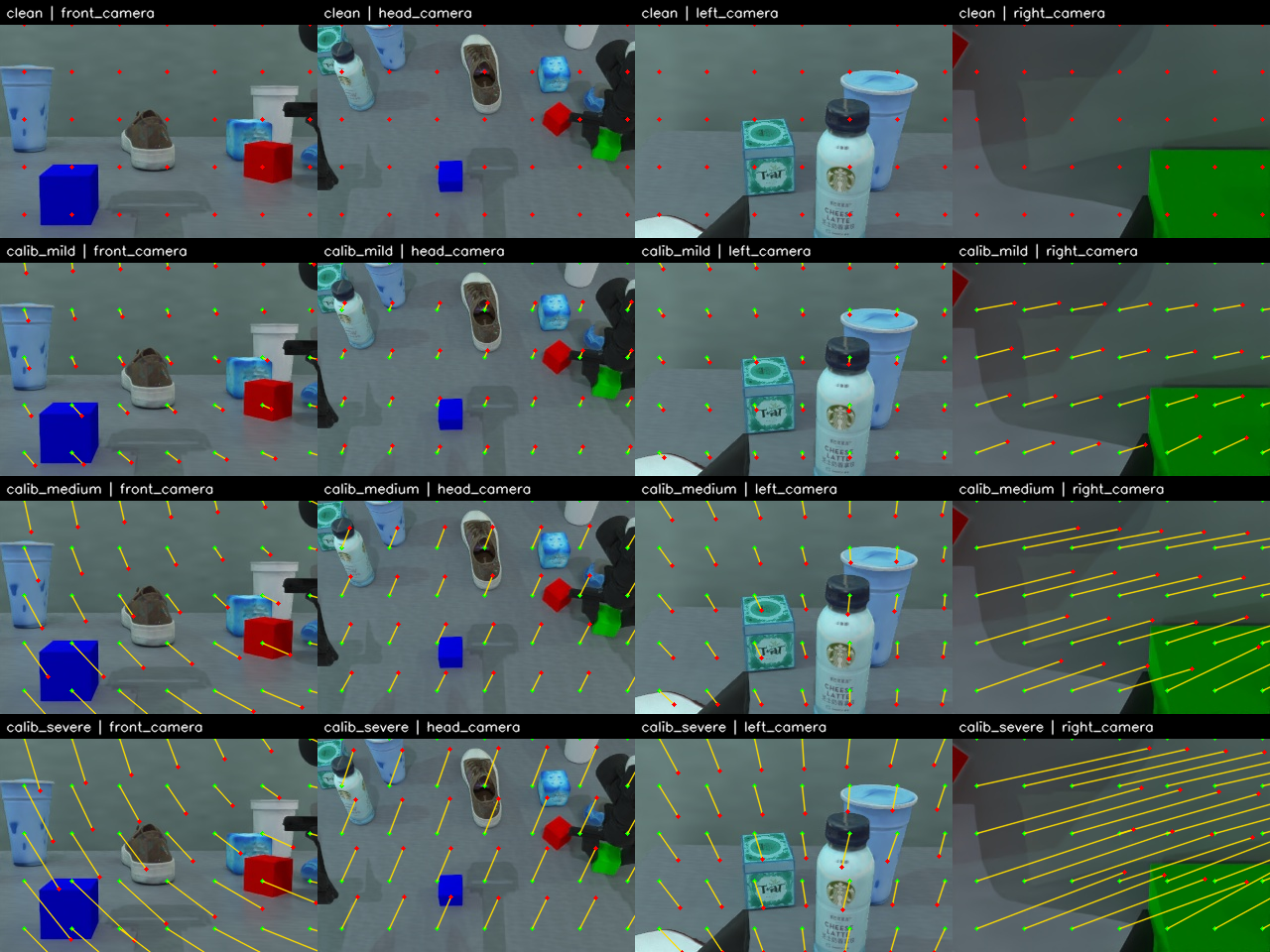}
    \caption{
    \textbf{Representative visualization of calibration perturbations.}
    Green and red markers denote original and perturbed projections, respectively, and yellow line segments show the induced image-space projection displacement. Each camera receives one episode-level perturbation that remains fixed throughout the episode, modeling persistent calibration bias rather than frame-wise jitter.
    }
    \label{fig:supp_calibration_projection_shift}
\end{figure}

\begin{figure}[t]
    \centering
    \includegraphics[width=0.85\linewidth]{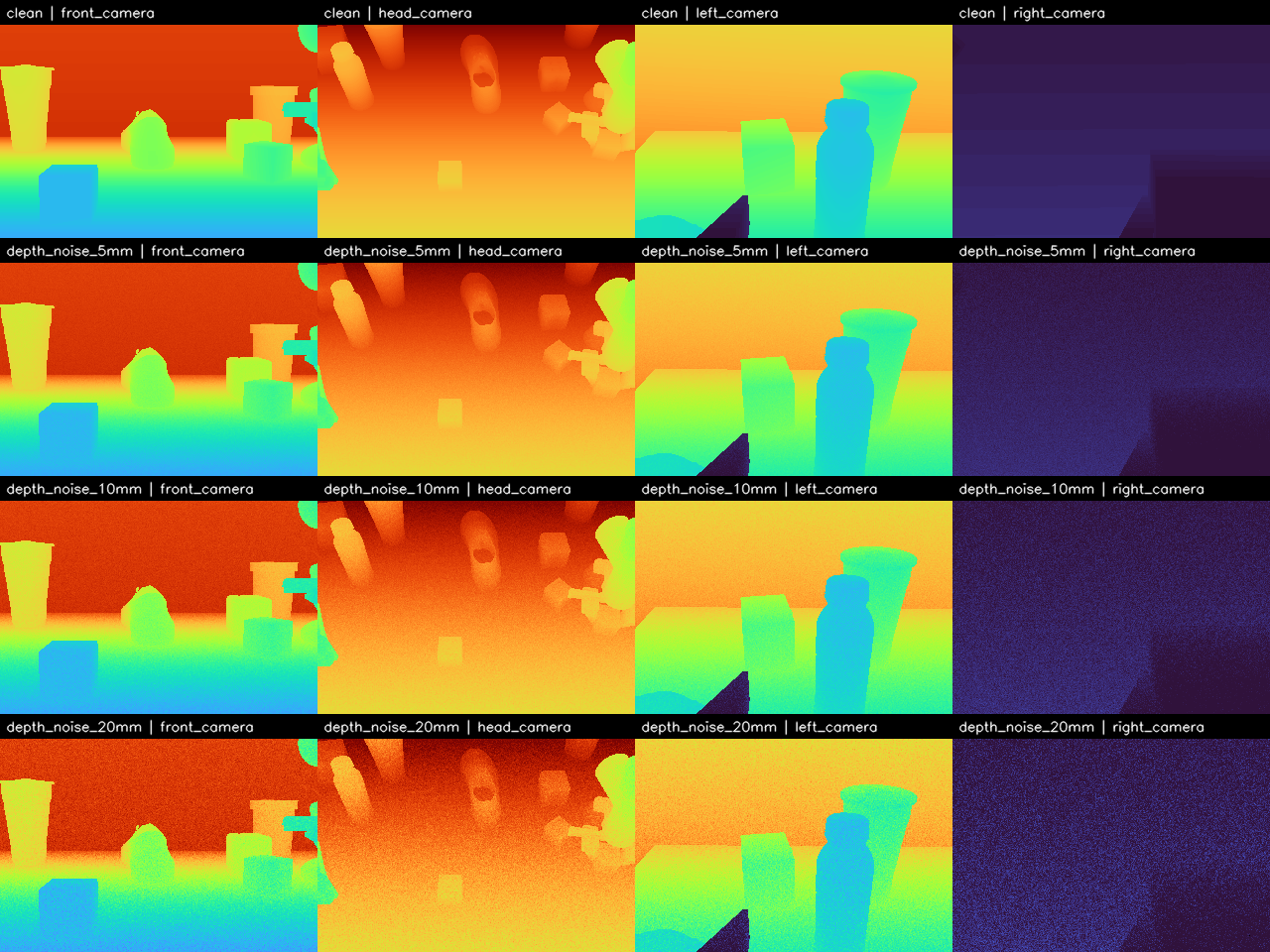}
    \caption{
    \textbf{Representative depth observations under additive Gaussian noise.}
    Rows from top to bottom correspond to depth-noise standard deviations of $0$, $5$, $10$, and $20$\,mm. Noise is added only to valid depth pixels, while originally invalid pixels remain invalid.
    }
    \label{fig:supp_depth_noise_visualization}
\end{figure}

\begin{figure}[t]
    \centering
    \includegraphics[width=0.85\linewidth]{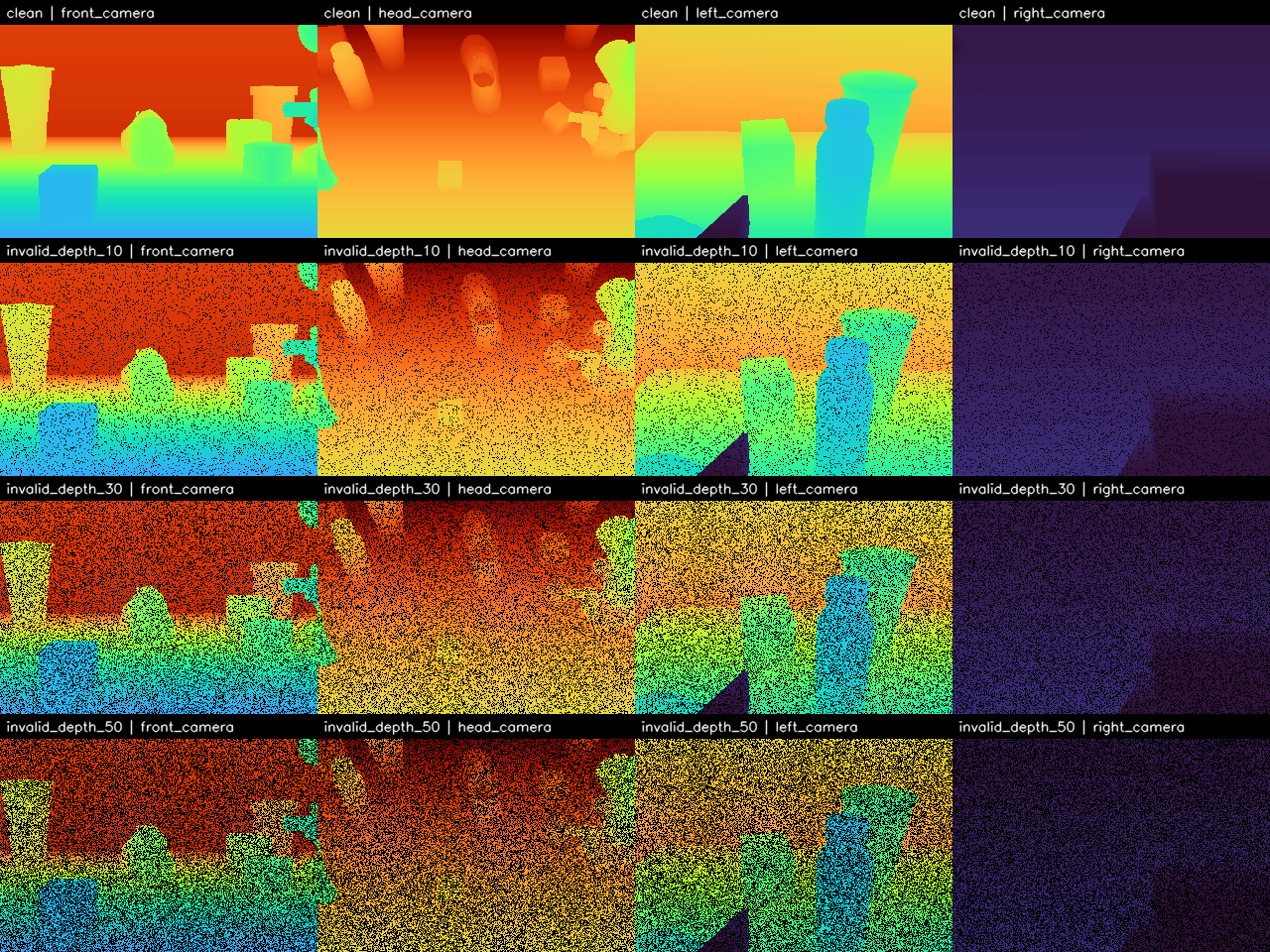}
    \caption{
    \textbf{Representative depth observations under missing-depth corruption.}
    Rows from top to bottom correspond to missing-depth ratios of $0\%$, $10\%$, $30\%$, and $50\%$. Missing-depth corruption randomly sets originally valid depth pixels to zero while leaving RGB observations and camera parameters unchanged.
    }
    \label{fig:supp_invalid_depth_visualization}
\end{figure}

\subsection{Robustness Results}
\label{app:robustness_results}

Table~\ref{tab:supp_geometry_robustness} reports results for the separately trained robustness-study checkpoints. Under matched no-perturbation settings, PEAfowl is 26.1 pp higher than the reproduced SEM reference on Clean and 26.5 pp higher on DR. The remaining rows evaluate each PEAfowl checkpoint under calibration and depth perturbations without retraining. Because these checkpoints use a different batch size and learning rate from the main-result checkpoints, their absolute success rates are interpreted only within this robustness-study configuration.

Within the evaluated ranges, additive depth noise produces smaller changes than severe missing-depth corruption or calibration error. This sensitivity is consistent with PEAfowl's use of expected 3D token anchors for cross-view retrieval, but the diagnostic experiment does not isolate a causal mechanism.

\section{Real-World Evaluation Protocol}
\label{app:real_world_protocol}

\subsection{Autonomous Trial Protocol}
\label{app:autonomous_protocol}

All formal real-world trials are fully autonomous. Once a trial starts, no human intervention is allowed until the trial terminates as either success or failure. Human intervention is used only between trials to reset the scene.

A trial is counted as a failure if any of the following occurs:
\begin{itemize}
    \item the robot requires manual correction during execution;
    \item the scene requires manual recovery before the trial terminates;
    \item the robot fails to complete the task within the 2-min execution horizon.
\end{itemize}

PEAfowl does not include an explicit failure detector or recovery planner. Since the policy is executed from current multi-view RGB-D observations, it may recover from some intermediate execution errors. For example, if a missed grasp leaves the object in a state that remains reachable and observable, the policy may reattempt the manipulation from the new observation. However, if the task is not completed within the allowed 2-min horizon, the trial is still counted as a failure.

\end{document}